\documentclass[letterpaper, 10 pt, conference]{ieeeconf}   
\usepackage{graphicx}
\usepackage{subcaption}
\usepackage{amsmath}
\usepackage{amssymb}
\usepackage{float}
\usepackage{url}

\graphicspath{ {images/} }

\newcommand{\nostarnote}[1]{}
\newcommand{\baad}[1]{} 

\IEEEoverridecommandlockouts                          

\title{\LARGE \bf
SafeDrive: Enhancing Lane Appearance for Autonomous and Assisted Driving Under Limited Visibility
}

\author{Jiawei Mo$^{1}$ and Junaed Sattar$^{2}$
\thanks{The authors are with the Department of Computer Science and Engineering, University of Minnesota Twin Cities, 200 Union St SE, Minneapolis, MN, 55455, USA
{\tt\small \{$^{1}$moxxx066, $^{2}$junaed\} at umn.edu.}}
}

\begin{document}

\maketitle
\thispagestyle{empty}
\pagestyle{empty}

\begin{abstract}

Autonomous detection of lane markers improves road safety, and purely visual tracking is desirable for  widespread vehicle compatibility and reducing sensor intrusion, cost, and energy consumption. However, visual approaches are often ineffective because of a number of factors; \emph{e.g.}, occlusion, poor weather conditions, and paint wear-off. We present an approach to enhance lane marker appearance for assisted and autonomous driving, particularly under poor visibility. Our method, named SafeDrive, attempts to improve visual lane detection approaches in drastically degraded visual conditions. SafeDrive finds lane markers in alternate imagery of the road at the vehicle's location and reconstructs a sparse 3D model of the surroundings. By estimating the geometric relationship between this 3D model and the current view, the lane markers are projected onto the visual scene; any lane detection algorithm can be subsequently used to detect lanes in the resulting image. SafeDrive does not require additional sensors other than vision and location data. We demonstrate the effectiveness of our approach on a number of test cases obtained from actual driving data recorded in urban settings.

\end{abstract}

\textbf{
}

%
\section{Introduction}
\label{sec:introduction}

Recent advances in affordable sensing and computing technologies have given new impetus towards commercialization of a wide variety of intelligent technologies. A major consumer-targeted application has focused on increasing autonomy in transportation systems, the most prominent of which is the area of self-driven cars. Autonomous driving has been a key focus in both academic and industrial research and development activities~\cite{Thrun2010TRC} as of late. Alongside fully autonomous commercial vehicles, mainstream auto manufacturers are equipping their vehicles with more intelligent technology with semi-autonomous, \emph{assistive} features -- the primary focus being increased safety. Many recent consumer-grade vehicles come with a number of such safety-enhancing features -- \emph{e.g.}, lane assist, blind-spot detection, radar-assisted braking, visual collision avoidance, driver fatigue detection~\cite{wang2006driver} -- with the number and quality of features increasing in higher-end and more expensive vehicles. These features are also available as add-on , albeit, expensive options. Even then, not all vehicles are capable of being fitted such a system, as these often require vehicle-specific data and power interfaces, limiting their application to newer vehicles. However, to minimize distracted driving (which has approximately 20 {\em per cent} contribution to fatalities on the road~\cite{distracteddata2012}) and improve safety, many consumers are opting to buy newer vehicles with these features pre-installed. 

\begin{figure}[t!]
    \vspace{2mm}
    \centering
    \begin{subfigure}[b]{0.22\textwidth}
        \includegraphics[width=\textwidth]{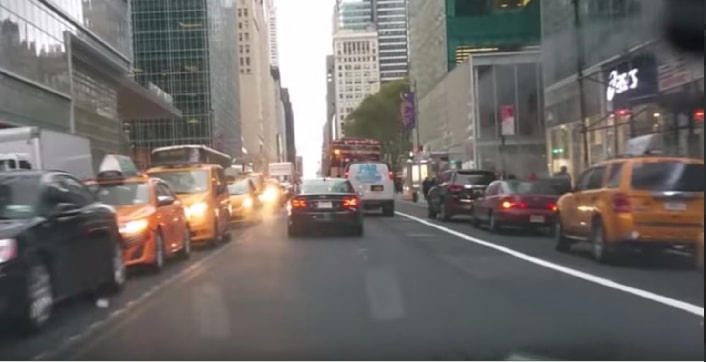}
        \caption{}
        \label{fig:Sequence_1}
    \end{subfigure}
    ~ 
    \begin{subfigure}[b]{0.22\textwidth}
        \includegraphics[width=\textwidth]{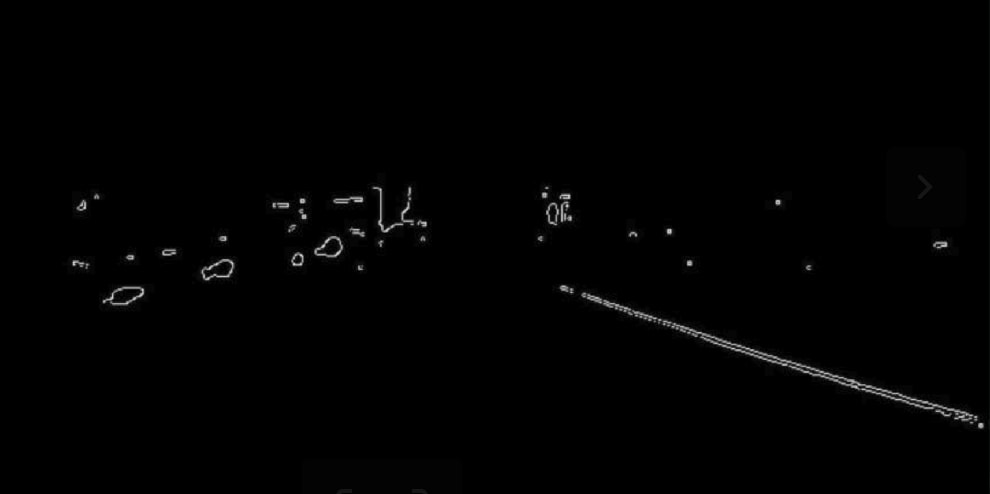}
        \caption{}
        \label{fig:Sequence_1_output}
    \end{subfigure}
    ~ 
    \begin{subfigure}[b]{0.22\textwidth}
        \includegraphics[width=\textwidth]{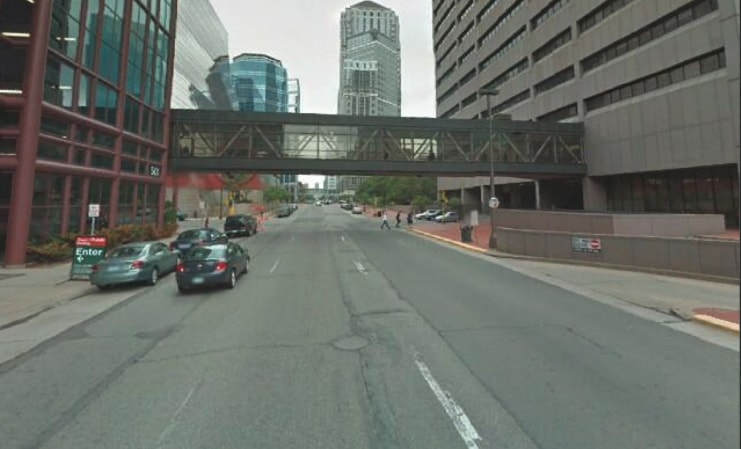}
        \caption{}
        \label{fig:Sequence_2}
    \end{subfigure}
    ~
    \begin{subfigure}[b]{0.22\textwidth}
        \includegraphics[width=\textwidth]{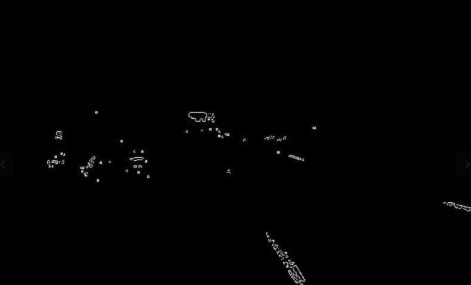}
        \caption{}
        \label{fig:Sequence_2_output}
    \end{subfigure}
    ~ 
    \begin{subfigure}[b]{0.22\textwidth}
        \includegraphics[width=\textwidth]{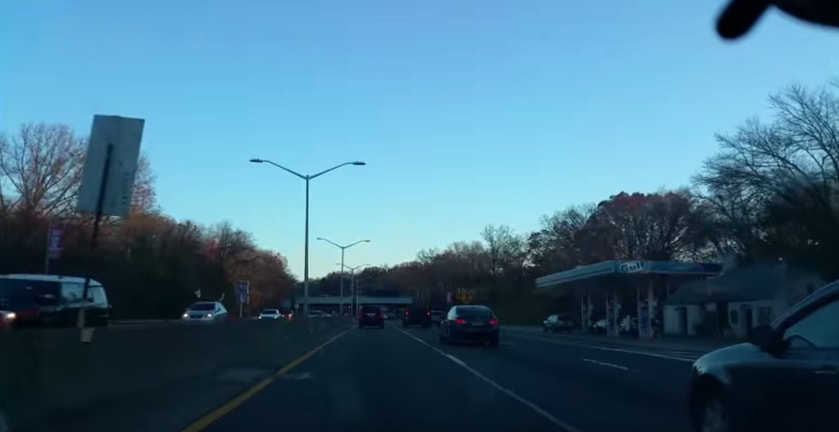}
        \caption{}
        \label{fig:Sequence_3}
    \end{subfigure}
    ~
	\begin{subfigure}[b]{0.22\textwidth}
        \includegraphics[width=\textwidth]{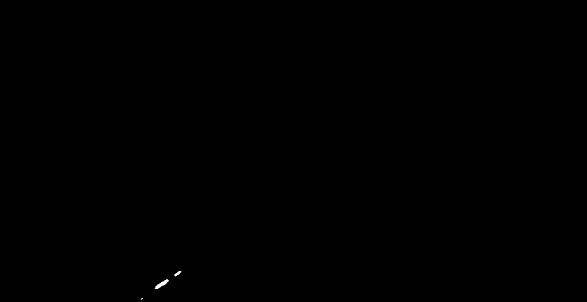}
        \caption{}
        \label{fig:Sequence_3_output}
    \end{subfigure}
    ~
    \begin{subfigure}[b]{0.22\textwidth}
        \includegraphics[width=\textwidth]{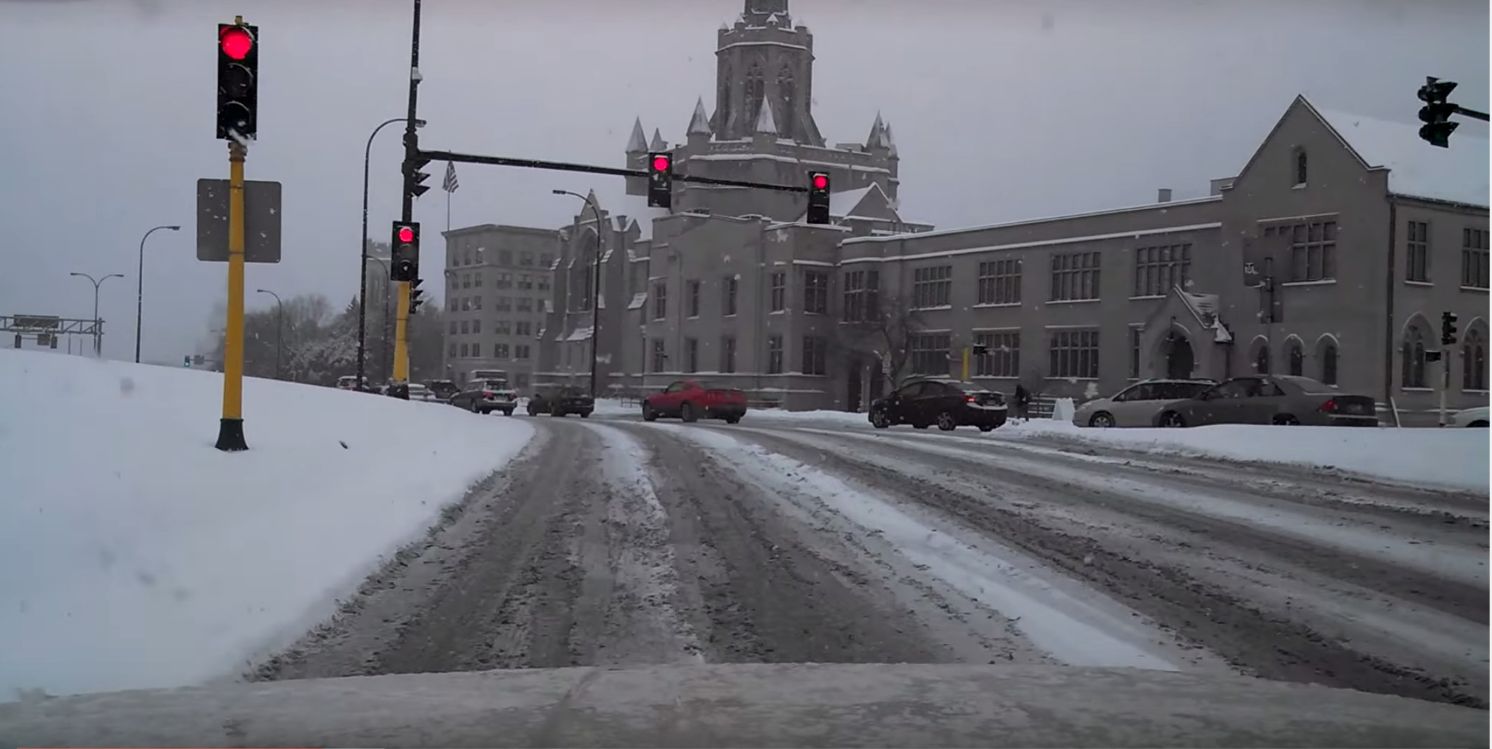}
        \caption{}
        \label{fig:Sequence_4}
    \end{subfigure}
    ~
    \begin{subfigure}[b]{0.22\textwidth}
        \includegraphics[width=\textwidth]{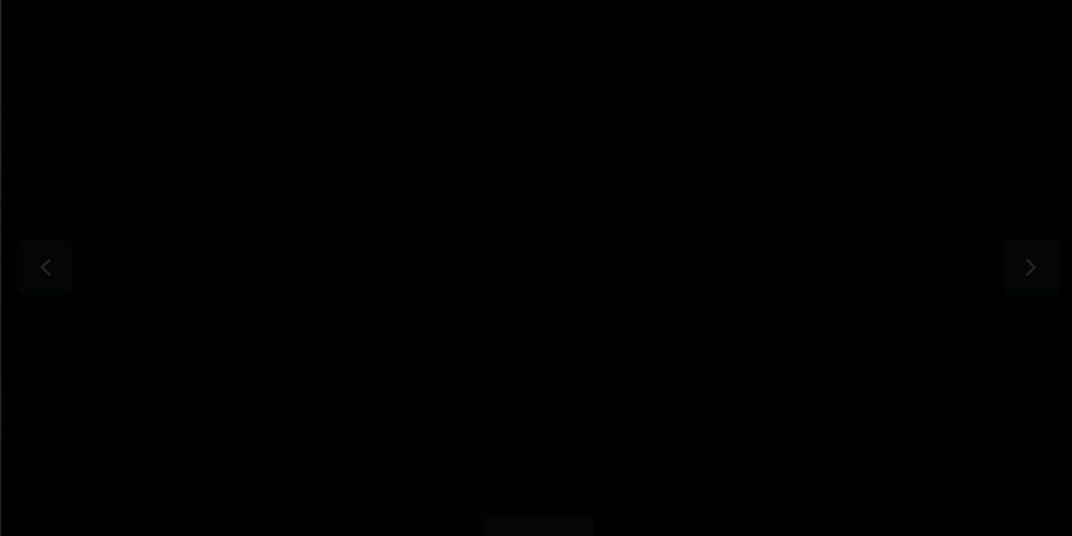}
        \caption{}
        \label{fig:Sequence_4_output}
    \end{subfigure}
    \caption{Visual lane tracking on several urban scenes from YouTube\texttrademark{} videos. Snapshot (\ref{fig:Sequence_1}) (output in (\ref{fig:Sequence_1_output})): lane markers not distinct in the center, though side markers are detectable. Snapshot (\ref{fig:Sequence_2}) (output in (\ref{fig:Sequence_2_output})): lane markers mostly washed out. Snapshot (\ref{fig:Sequence_3}) (output in (\ref{fig:Sequence_3_output})): evening drive, low-light conditions make the lane markers almost undetectable. Snapshot (\ref{fig:Sequence_4}) (output in (\ref{fig:Sequence_4_output})): snow-covered roads, no lane markers detected.}
\label{fig:Urban_Sequence}
\end{figure}

Across manufacturers (and in some cases, vehicle models), a variety of sensing methods are used to provide accurate detection of road features and consequently prevent traffic mishaps. Such sensors include but are not limited to laser scanners, radar, proximity sensors, and visible-spectrum cameras. Vision is an unobtrusive, low-cost, low-power sensor but requires appropriate lighting, unobstructed views, and fast processing times for deployment in autonomous and assisted driving applications. Nevertheless, vision sensors carry a strong appeal for deployment in mass production systems, mostly because of its low-power, inexpensive nature. For example, Mobileye~\cite{mobileye2004forward} is a well-known monocular visual system used for lane tracking, collision avoidance and vehicle safety, and is typically deployed as an add-on hardware components in existing vehicles. Subaru also uses a stereo vision system called EyeSight~\cite{SubaruEyesight} for lane detection and collision avoidance, where a stereo camera pair is mounted on either side of the center rear-view mirror. This provides depth perception and lane tracking, and an intelligent drive system provides adaptive braking and cruise control, collision avoidance and lane-departure warnings. While such systems have shown to work reasonably well, they are not immune to failures arising from degraded visual conditions. A few example scenarios are shown in Figure~\ref{fig:Urban_Sequence}, which demonstrate how changing conditions affect the quality of the center lane markers in the visual scene. The lane detection system in all four cases is using a real-time segmentation approach; however, irrespective of the particular algorithm used, the input images themselves are of significantly degraded quality for robust lane tracking.

This paper proposes a system called \textbf{SafeDrive}, which is a significantly inexpensive approach for enhancing visual lane appearance in severely degraded conditions, without relying on exotic, costly sensors which would be prohibitive for financial and compatibility reasons. Under poor visibility, the system uses vehicle's location data to locate alternate images of the road from an available ``road-view'' database. A sparse 3D street model is reconstructed from the alternate images. Subsequently, lane markers in the 3D model are projected onto current view, according to the geometric relationship between current view and 3D street model. \emph{SafeDrive is designed to be used as a preprocessing step for visual lane tracking systems in adverse conditions, and not a lane tracking system by itself}. For development of SafeDrive, an Android-based application called DriveData has been developed to capture a variety of data from a device mounted on (or even outside) the vehicle. Our long-term goal is to create an affordable solution (\emph{e.g.}, to be used on a smartphone mounted on the windshield) to provide lane departure warnings in extreme cases, and also provide safety recommendations under the current driving conditions based on visual, location and acceleration data.

\paragraph*{\textbf{Additional Related Work}}
\label{sec:related}

A large body of literature exists on different aspects of autonomous driving and driver's assistance technologies, a number of which relies on robotics and computer vision methods. The Carnegie-Mellon NavLab~\cite{Thorpe_1985_2467} project has produced some of the earliest implementations of self-driving cars, and have extensively used vision for a number of subtasks, including road and lane detection~\cite{kluge1990explicit}. Stereo vision systems have been used for lane detection; \emph{e.g.}, in~\cite{nedevschi20043d,bertozzi1998gold}. Dedicated parallel processing for road and lane detection have been investigated in the GOLD~\cite{bertozzi1998gold} framework. Kluge et al.~\cite{kluge1995deformable} applied deformable shapes and Wang et al.~\cite{wang2004lane} used ``snakes'' for detection and tracking of lane markers. Spline-fitting methods for lane tracking have also been applied~\cite{wang1998lane}. Lane markers throughout the world are often painted yellow or white, which has led a number of researchers to investigate color-based approaches for finding lane makers (\emph{e.g.},~\cite{chiu2005lane,gonzalez2000lane,borkar2009robust}). Often used with a combination Bayesian filtering and estimation methods, these methods have shown to work well under clear visibility conditions. Some researchers looked into the problem of rear-view lane detection~\cite{takahashi2002rear}. Abramov~\cite{abramov2016multi} looks at lane detection using stereo cameras resulting in a system similar to the Subaru EyeSight\texttrademark mechanism. Advances in machine vision with deep learning~\cite{goodfellow2016deep} methods have been used in more recent work on autonomous driving, particularly in artifact and lane detections (\emph{e.g.},~\cite{john2014traffic}). The paper by Hillel et al.~\cite{hillel2014recent} is an in-depth survey of recent advances in road and lane detection problems. Kim~\cite{kim2008robust} investigated robust lane tracking under challenging conditions with poor visibility and rapidly changing road traffic. Son et al. proposes an illumination invariant lane detection method in~\cite{son2015real}. However, these methods do not consider cases where visibility of lane markers are extremely degraded, to the point of becoming invisible.
\section{Methodology}
\label{sec:methodology}

\begin{figure}[tp]
	\centering
    \includegraphics[width=0.48\textwidth]{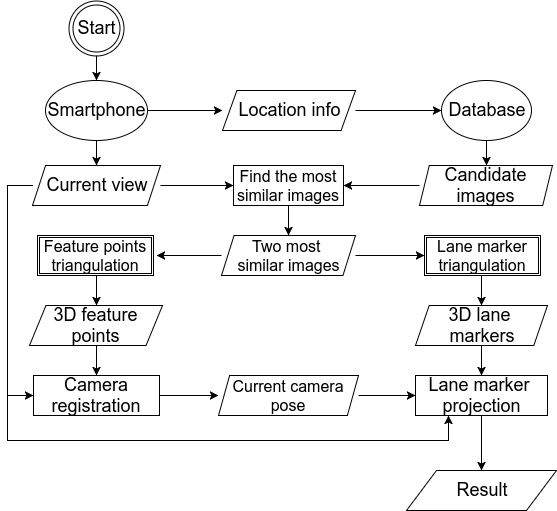}
    \caption{SafeDrive operational flowchart.}
    \label{fig:flowchart_overall}
\end{figure}

SafeDrive is based on the assumption that \emph{alternate images} of the current location are available and are of better visual quality; therefore, lane markers can be found in those images and subsequently projected onto the current, degraded image. In SafeDrive, lanes are projected using geometric correspondences between the current view and a sparse reconstructed view of the current scene obtained from alternate imagery, using the Structure from Motion\cite{Hartley2004} approach. A database of images, indexed by location, is searched to find images most similar to the current scene, and a 3D model of the street is reconstructed from these images. The current view is then registered with this 3D model, and the lane markers in the 3D model are projected onto the current view (see Figure~\ref{fig:flowchart_overall}).

\subsection{Finding the most similar images}
\label{sec:find_images}

In order to get the alternate images for 3D reconstruction, we search a database for images most similar to the current scene being observed by the camera. These images are referred to \emph{candidate images}, while the actual camera image is referred to as the \emph{current image}. Images in the database are indexed by their location so that only a small subset of images that were taken near the current location are searched. For our purposes, the criterion we use to measure \emph{similarity} between two images is \emph{simply the number of matched feature points}. While this may seem overly simplistic, particularly as the current and candidate images may have very different appearances, this is sufficient as our intent is to find two images with maximally overlapped visual content. This requirement essentially ensures the highest number of features match for realistic scenes. Feature points are detected using the Harris corner detector\cite{harris1988combined}. Instead of keeping every feature point, we only keep those which are at least a minimum distance away from any other feature point. This is to ensure a maximal spatial distribution of features (\emph{i.e.}, between the left and right side of the image) to improve the likelihood of finding the most accurate image match. The ORB (Oriented and Rotated BRIEF) feature descriptor~\cite{rublee2011orb} is used to extract and match features from both the current and the candidate images. To ensure an accurate match, descriptor matching is run in both direction between these two images to further remove inconsistent matches.

After matching features on all possible candidate images at the current location, a minimum of two images with the most number of matched points are chosen for 3D reconstruction. A higher number of images improves the quality of the 3D street model; however, as our focus is not on the 3D reconstruction itself, we use two images to reduce the computation load. While such a choice may question the resulting accuracy of the entire process, our experiments demonstrate that using two images for the reconstruction process is sufficient for SafeDrive if these images are taken from locations moderately distant (\emph{i.e.}, neither too far nor too close) from each other. 

\subsection{Feature Points Triangulation}
\label{sec:feature_triangulation}
\begin{figure}
	\centering
    \includegraphics[width=0.48\textwidth]{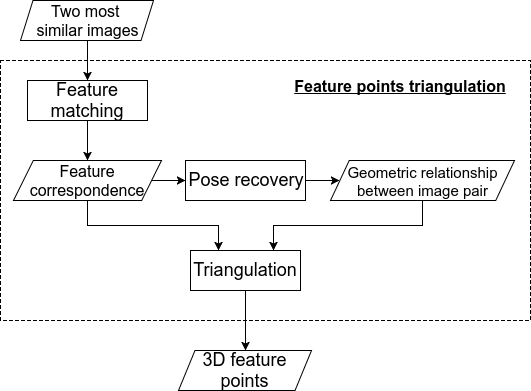}
    \caption{Diagram depicting feature points triangulation in SafeDrive.}
    \label{fig:flowchart_feature}
\end{figure}

Since we are only interested in the geometric relationship between the 3D street model and current view, instead of full 3D reconstruction, we build a sparse 3D street model with feature points only, which is sufficient for camera registration (see Figure~\ref{fig:flowchart_feature}). To generate the 3D points needed for sparse 3D street model reconstruction, we need to estimate the relative pose between the two images. We match features between the image pair, then the correspondences are used to calculate the fundamental matrix between the two images. Subsequently, the rotation matrix and translation vector between the image pair are recovered from the fundamental matrix and camera intrinsic parameters. The translation vector is up to scale, but that does not pose a problem. To understand why, assume the scale is $\lambda$, and the generated 3D point is $\lambda (x,y,z),$ where $(x,y,z)$ is the position of this point in world coordinates. When registering the current view, the current camera pose will be $[\textbf{R}, \lambda \textbf{t}] \in \mathbb{SE}(3)$, compared to the real current camera pose $[\textbf{R}, \textbf{t}] \in \mathbb{SE}(3)$, where $\textbf{R} \in \mathbb{SO}(3)$ is the rotation matrix and $\textbf{t} \in \mathbb{R}^3$ is the translation vector. As a result, the transformed point in the current camera frame before projection will be: $[\textbf{R}, \lambda \textbf{t}][\lambda x, \lambda y, \lambda z, 1]^T = \lambda [\textbf{R}, \textbf{t}][x, y, z, 1]^T$. Since we are using the pinhole camera model, the scale factor is eventually eliminated.

Based on the rotation matrix and translation vector, 2D feature correspondences are triangulated to get 3D points. To solve for each 3D feature point $X$, we use $K$ to represent the camera intrinsic parameters, and $[R_i, t_i]$ to represent the pose of the $i$th camera in world coordinate. For feature $u_1$ in the first image:
\begin{align*}
	\lambda_1 u_1 = K[R_1, t_1]X \\
    u_1 \times \lambda u_1 = u_1 \times K[R_1, t_1]X \\
    0 = [u_1 \times K[R_1, t_1]] X
\end{align*}
Similarly for its corresponding feature $u_2$ in the second image:
\begin{align*}
    0 = [u_2 \times K[R_2, t_2]] X
\end{align*}
By stacking them together, we have 
\begin{align*}
	\begin{bmatrix}
		u_1 \times K[R_1, t_1] \\
		u_2 \times K[R_2, t_2]
	\end{bmatrix}
    X = 0
\end{align*}
To obtain X, we apply Singular Value Decomposition (SVD) on 
\begin{align*}
\begin{bmatrix}
	u_1 \times K[R_1, t_1] \\
	u_2 \times K[R_2, t_2]
\end{bmatrix}
\end{align*}

As lane markers are lines, it is not possible to match them directly between a pair of images using feature points, and as a result, their locations cannot be triangulated in similar manner. Our solution is thus to first detect pixels which belong to lane markers on both images, then match them between the rectified image pair.

\subsection{Lane Marker Triangulation}
\label{sec:marker_triangulation}

\begin{figure}
	\centering
    \includegraphics[width=0.48\textwidth]{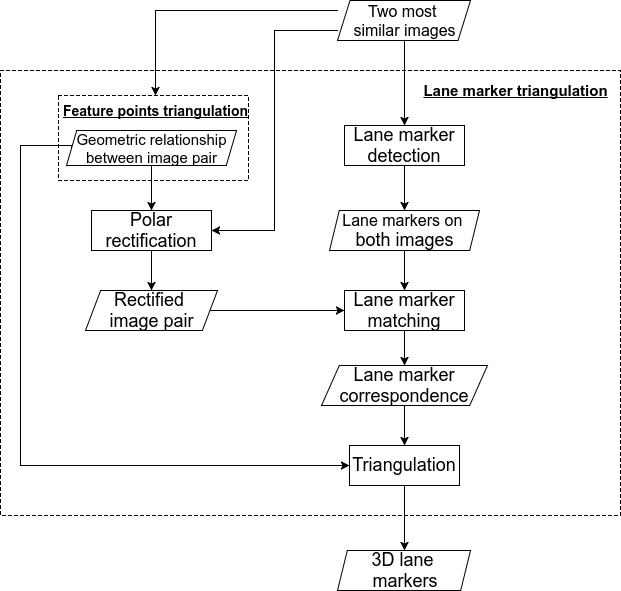}
    \caption{Diagram depicting lane marker triangulation in SafeDrive.}
    \label{fig:flowchart_lane}
\end{figure}
\baad{The strategy of lane marker detection is based on the color histogram and Canny edge detection algorithm.}For each image, we extract yellow or white pixels based on their pixel value, since most lane markers in U.S. traffic system are of these two colors. Although this is performed on the candidate images where lanes are clearly visible, we convert image from RGB to HSV color space to make it more robust to illumination changes. At the same time, we run Canny edge detection~\cite{canny1986computational} to extract the edges on the image. Finally, we take the intersection of yellow/white pixels and edges on the image as locations of lane markers (see Figure~\ref{fig:flowchart_lane}). \baad{From experiments, we find this method is robust enough for our project.}

\begin{figure}
	\centering
    \includegraphics[width=0.48\textwidth]{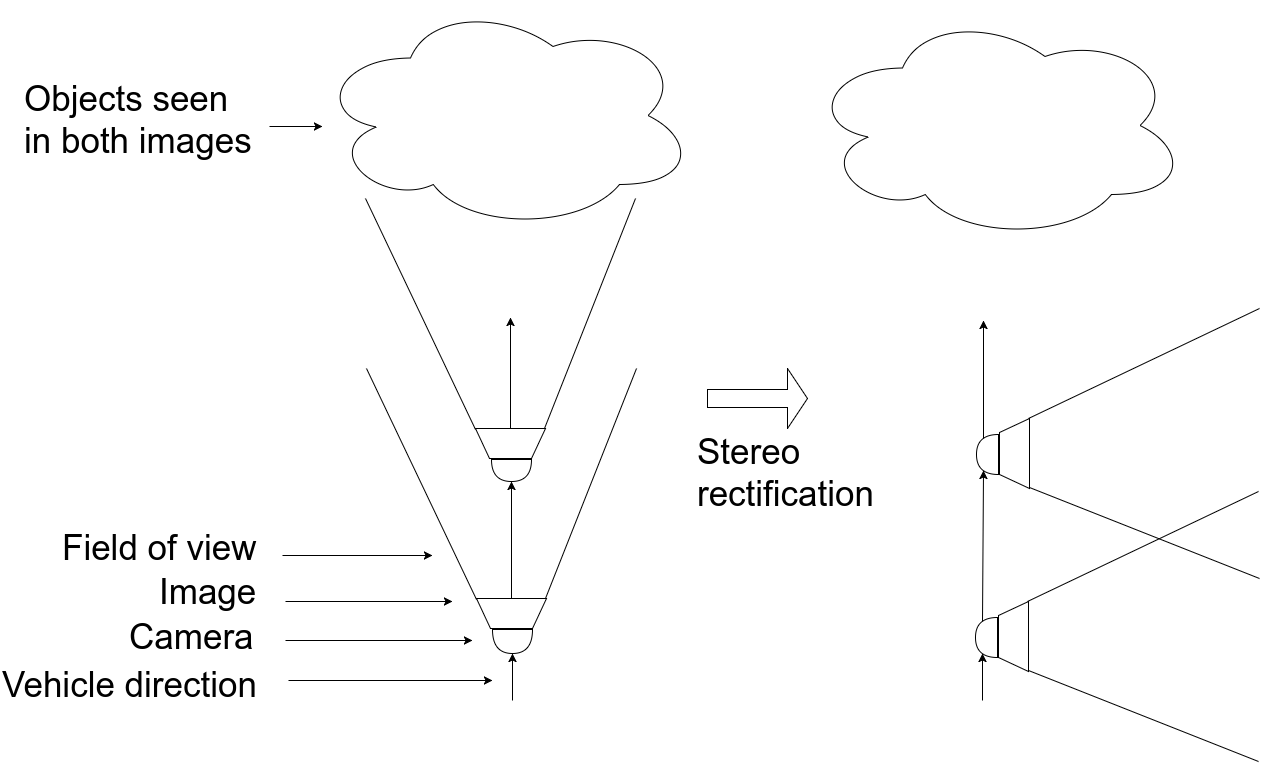}
    \caption{Mimicking stereo rectification with a forward-moving camera.}
    \label{fig:stereo_rect}
\end{figure}

We use the fundamental matrix computed in the previous step to rectify the image pair for triangulation. One important observation here is that vehicles are predominantly forward-moving; thus, if we use classic stereo rectification~\cite{papadimitriou1996epipolar}, most of the information on the original image will disappear from the rectified image because camera is rotated almost $90^{\circ}$, as illustrated in Figure~\ref{fig:stereo_rect}. To prevent this, we rectify the image pair in polar coordinates around the epipole~\cite{pollefeys1999simple}. The vertical axis of the rectified image is the polar angle around the epipole, and the horizontal axis is the distance to the epipole. Polar rectification preserves pixels from the original image for camera movement in any arbitrary direction.

In order to match lane marker pixels on the rectified image, we first transform these lane marker pixels from the image pair to polar coordinates based on polar rectification. Consequently, each lane marker pixel from one image should have the same vertical coordinate as its corresponding pixel from the other image, because they share the same angle with respect to the epipole. As a result, search space is reduced to one dimension only, which significantly speeds up the searching process. Furthermore, we do not need to check every pixel along the horizontal line, but check only those lane maker pixels that are likely to be the correspondence, which reduces the search space further\baad{ to several pixels}. Currently, we compute a descriptor for each pixel, and compare it with the descriptors of those lane marker pixels in the other image that have the same vertical (\emph{i.e.}, Y) coordinates.\baad{ It works fine but not very robust. Future work should stress on better lane maker matching.} After getting lane marker pixel correspondence on the polar coordinates, we transform the pixels back to Cartesian coordinates, and triangulate them using the methods described in Section~\ref{sec:feature_triangulation}. To further remove outliers, we reproject the triangulated 3D road marker to the original image pair and remove those road makers whose reprojection error exceeds a preset threshold. 

\subsection{Camera Registration}
\label{sec:camera_registration}

After building a 3D street model that consists of feature points and lane marker points, the next step is to register the current view with respect to the reconstructed 3D street model. We take the descriptors of each feature points from the database images, and match them against the feature descriptors from current view. After that, we have 3D to 2D correspondence, based on which, we solve the Perspective-N-Point (PNP) problem~\cite{Hartley2004} to get the relative pose of current camera with respect to the 3D model. We multiply the camera intrinsic parameters with the relative pose to get the projection matrix. Finally, we are able to project the lane markers from the 3D model onto current view.\baad{, so that driver is able to see the projected lane markers through the smartphone.}\nostarnote{we need to say up front why this projection is needed -- maybe we already did, but need to remind the reader about this}

\section{Experimental Evaluation}
\label{sec:experiments}

\begin{figure*}[t!]
	\centering
	\includegraphics[width=\textwidth]{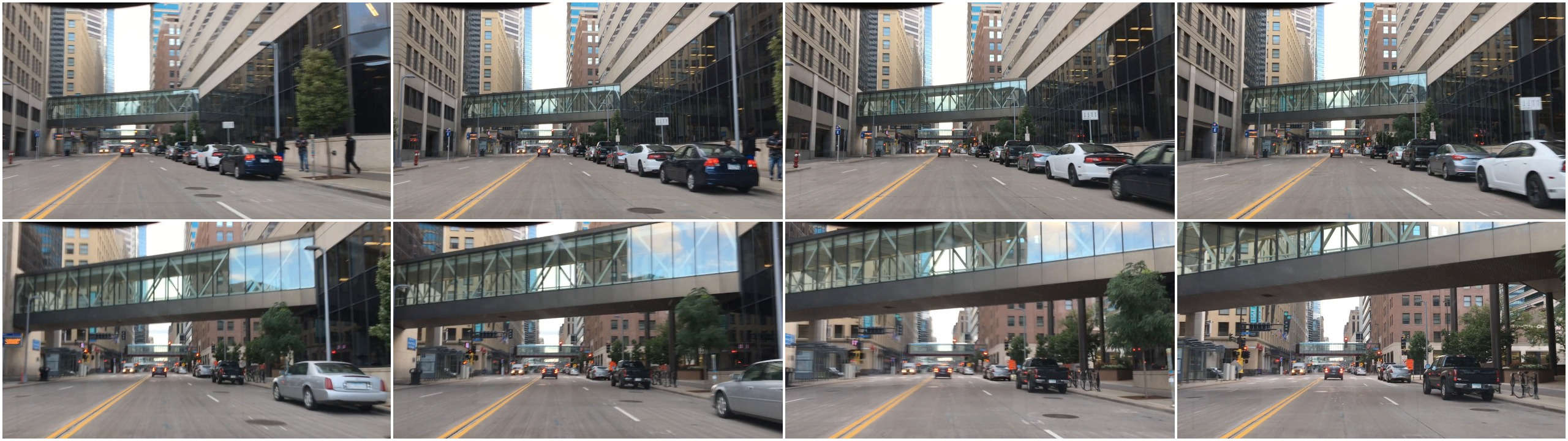}
	\caption{Images in the database around the location $(44.979238N, 93.266568W)$.}
	\label{fig:database}
\end{figure*}

We have evaluated the performance and accuracy of SafeDrive using driving data acquired from a windshield-mounted smartphone. We recorded two driving datasets, one of which served as the database for alternate imagery, with its lane markers visible. Current view of the road comes from the other dataset, which is recorded at different times on different days under different driving (and resultantly, road) conditions. The current view is artificially corrupted by hand to simulate poor visibility of lane markers. Additionally, in the current view, direct lane detection is not attempted at all; the detected lanes are projected onto this at the end.

\begin{figure}[h!]
  \centering
    \includegraphics[width=0.48\textwidth]{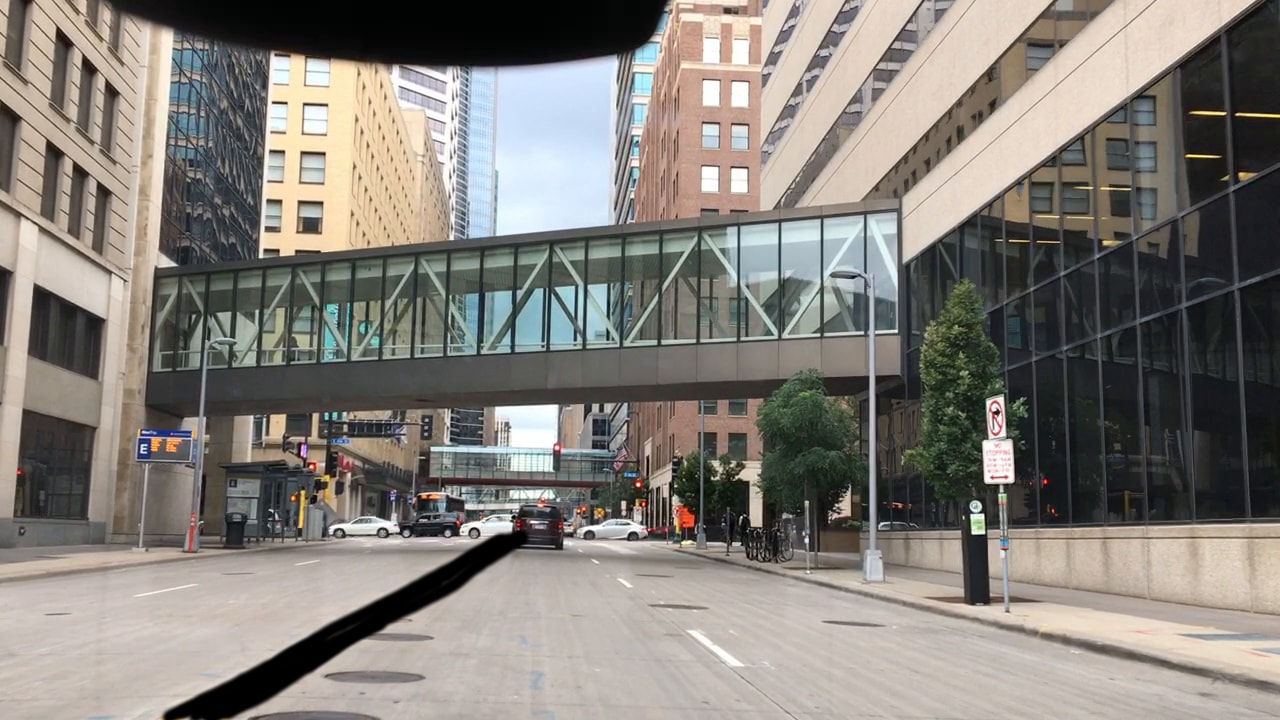}
   \caption{Current view around $(44.979238N, 93.266568W)$. Note that the center lane markers have been removed by hand from the current view for evaluation.}
   \label{fig:current}
\end{figure}

The test case is located around coordinates $(44.979238N, 93.266568W)$. We have eight images in the database for that location, as illustrated in Figure~\ref{fig:database}. The current view is selected from the other dataset around the same place, as shown in Figure~\ref{fig:current}. Note that the database can be arbitrarily large, and there are no restrictions on how many candidate images can be stored for any given location.

\begin{figure}[h!]
  \centering
  \begin{subfigure}[t]{0.48\textwidth}
    \centering
    \includegraphics[width=\textwidth]{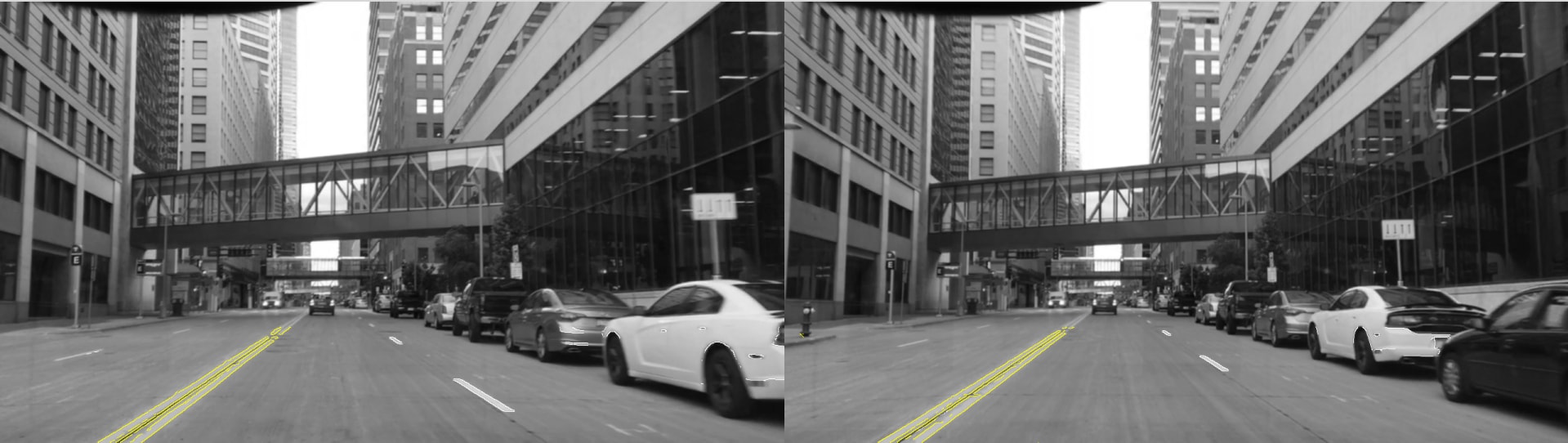}
   \caption{The most similar image pair in the database found by feature-based matching, and lane markers detected shown in yellow.}
   \label{fig:lane_markers}
  \end{subfigure}
    ~ 
  \begin{subfigure}[t]{0.48\textwidth}
    \centering
    \includegraphics[width=\textwidth]{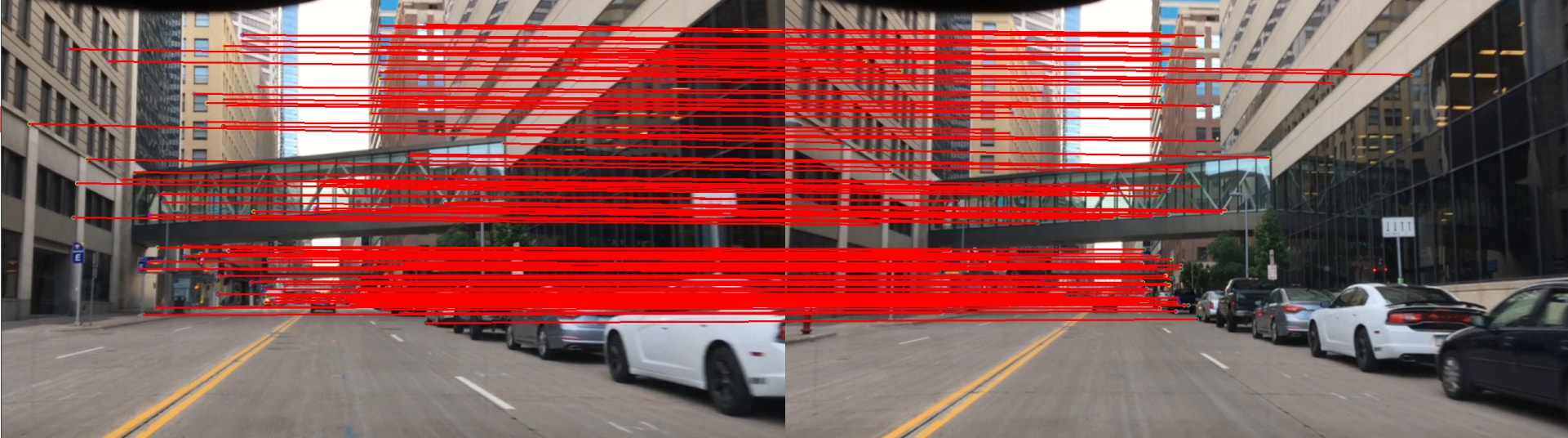}
   \caption{Feature correspondences between the most similar image pair.}
   \label{fig:feature_match}
  \end{subfigure}
    ~ 
  \begin{subfigure}[t]{0.48\textwidth}
    \centering
    \includegraphics[width=\textwidth]{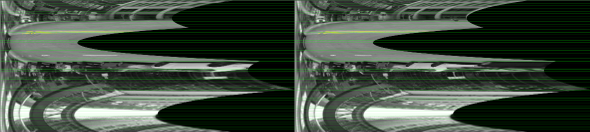}
    \caption{Rectified image pair in polar coordinates, with height reduced to 1/16 of original size (897x3204) for display purposes.}
    \label{fig:rectified}
   \end{subfigure}
    ~ 
  \begin{subfigure}[t]{0.48\textwidth}
    \centering
    \includegraphics[width=\textwidth]{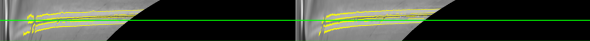}
    \caption{Lane marker in the rectified image pair.}
    \label{fig:rect_lane_match}
   \end{subfigure}
    ~ 
  \begin{subfigure}[t]{0.48\textwidth}
    \centering
    \includegraphics[width=\textwidth]{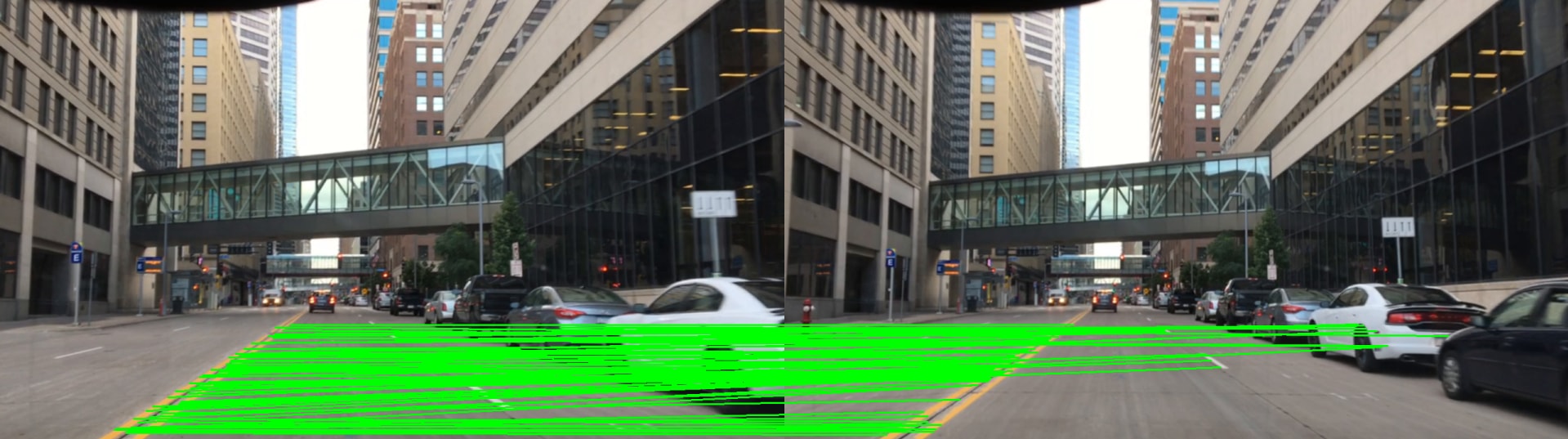}
    \caption{Lane marker correspondences between the images of the chosen pair.}
    \label{fig:lane_marker_match}
   \end{subfigure}
   \caption{The process of finding feature points and lane marker correspondences.}
\end{figure}

By matching features between the current view and all database images, the two images with the most feature matches are selected as the most similar image pair, which are shown in Figure~\ref{fig:lane_markers}. Afterwards, lane markers are extracted from both images for later use (as described in Section~\ref{sec:marker_triangulation}. We run feature matching between the images of this image pair (see Figure~\ref{fig:feature_match}) to calculate the fundamental matrix. For this test case, the fundamental matrix takes the following form:
\begin{equation*}
  \begin{bmatrix}
  3.7989e-07 & -0.0005 & 0.2287 \\
   0.0005 & 1.3512e-06 & -0.2502 \\
   -0.2294 & 0.2476 & 1.0
  \end{bmatrix}
\end{equation*}
The fundamental matrix is further used to rectify the image pair. The rectified image pair is shown in Figure~\ref{fig:rectified}. In this test case, since the vehicle was moving parallel to the lane marker, the lane marker is almost horizontal in the rectified image, as illustrated in Figure~\ref{fig:rect_lane_match}. This makes the matching process more challenging because there are many possible correspondences along the horizontal line. After matching lane marker pixels horizontally between the rectified image pair, as described in ~\ref{sec:marker_triangulation}, we get the lane marker correspondence as shown in Figure~\ref{fig:lane_marker_match}. The outliers in Figure~\ref{fig:lane_marker_match} are removed later by reprojection.

Besides rectification, the fundamental matrix is also used to recover the relative pose between the image pair. The intrinsic parameter of camera we use is:
\begin{equation*}
  \begin{bmatrix}
  1261.46807 & 0.0 & 619.89385 \\
  0.0 & 1259.44016 & 356.46599 \\
  0.0 & 0.0 &1.0
  \end{bmatrix}
\end{equation*}
Consequently, the recovered rotation matrix is:
\begin{equation*}
  \begin{bmatrix}
    0.999996 & 0.002211 & -0.001646 \\
     -0.002212 & 0.999997 & -0.000380 \\
     0.001645 & 0.000383 & 0.999999
  \end{bmatrix}
\end{equation*}
and translation vector is:
\begin{equation*}
  \begin{bmatrix}
    -0.069823 \\
     0.096280 \\
     0.992902
  \end{bmatrix}
\end{equation*}
As can be seen, the rotation matrix is close to identity, and translation vector is dominated by the Z-axis, as the vehicle was driving forward at that time. This corroborates the assumption that road-going vehicles do not usually have large lateral displacements when driving.

\begin{figure}[h!]
  \centering
  \begin{subfigure}[t]{0.48\textwidth}
    \centering
    \includegraphics[width=\textwidth]{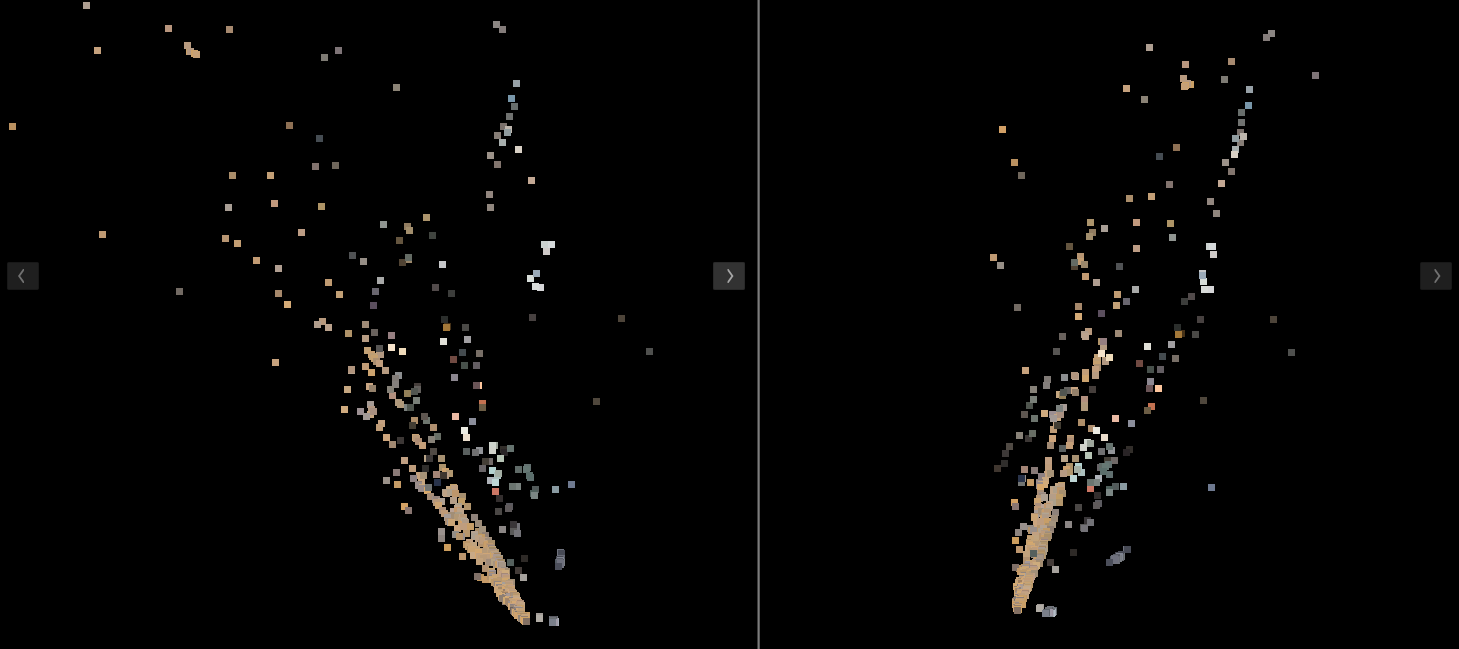}
    \caption{Reconstructed sparse 3D model of the street.}
    \label{fig:reconstructed_model}
   \end{subfigure}
    ~ 
  \begin{subfigure}[t]{0.48\textwidth}
    \centering
    \includegraphics[width=\textwidth]{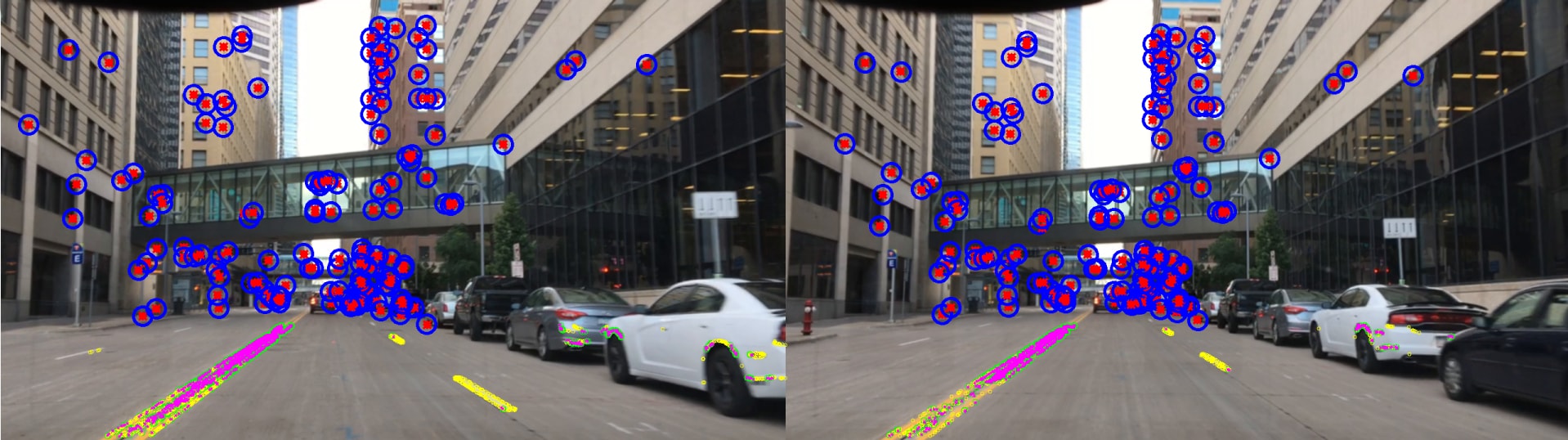}
    \caption{Reprojecting 3D points onto the image pair. Blue circles: matched feature point; red markers: reprojected 3D feature point; green circles: matched lane marker; magenta markers: reprojected lane marker; yellow circles: lane marker with larger reprojection error.}
    \label{fig:stereo_reproj}
   \end{subfigure}
   \caption{The final reconstructed 3D street model.}
\end{figure}

The rotation matrix and translation vector are used to triangulate both feature correspondences in Figure~\ref{fig:feature_match} and lane marker correspondences in Figure~\ref{fig:lane_marker_match}. The result of sparse street reconstruction by triangulation is shown in Figure~\ref{fig:reconstructed_model}. In order to measure the quality of both triangulations, we reproject the triangulated 3D points onto the original image pair, and measure their distances to the 2D correspondence. The average reprojection error of feature points is $0.368289$ pixels, and the average reprojection error of lane marker pixels is $0.897794$ pixels, which does not include outliers whose reprojection errors exceed a preset threshold, and are marked with yellow circles in Figure~\ref{fig:stereo_reproj}.
  
\begin{figure}[h!]
  \centering
  \begin{subfigure}[t]{0.48\textwidth}
    \centering
    \includegraphics[width=\textwidth]{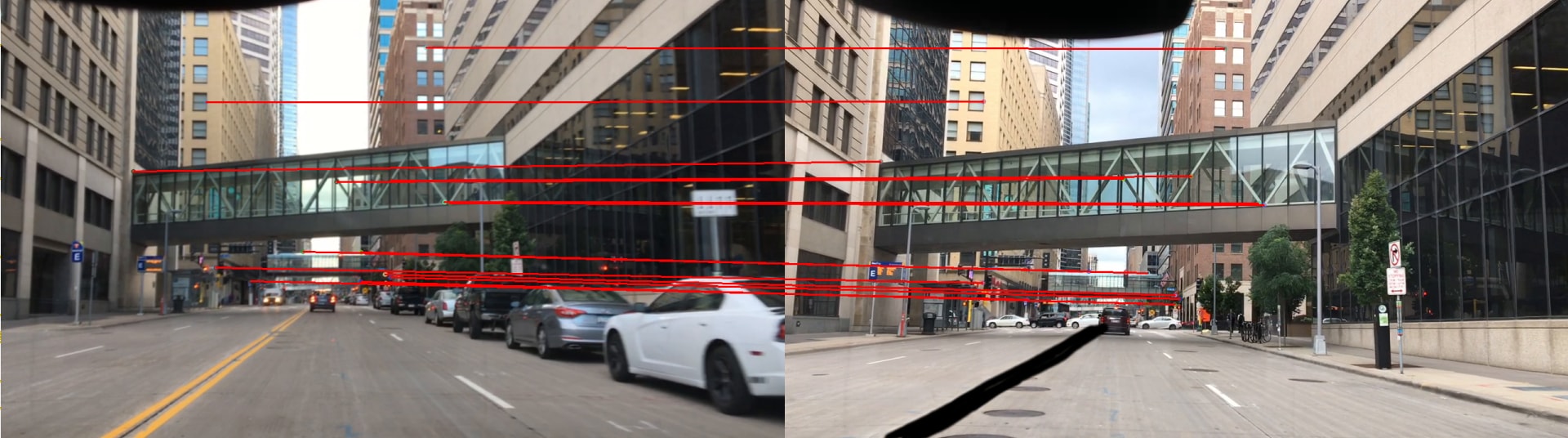}
   \caption{Feature match between the current view and database image.}
   \label{fig:PnP_inlier}
  \end{subfigure}
    ~ 
  \begin{subfigure}[t]{0.48\textwidth}
    \centering
    \includegraphics[width=\textwidth]{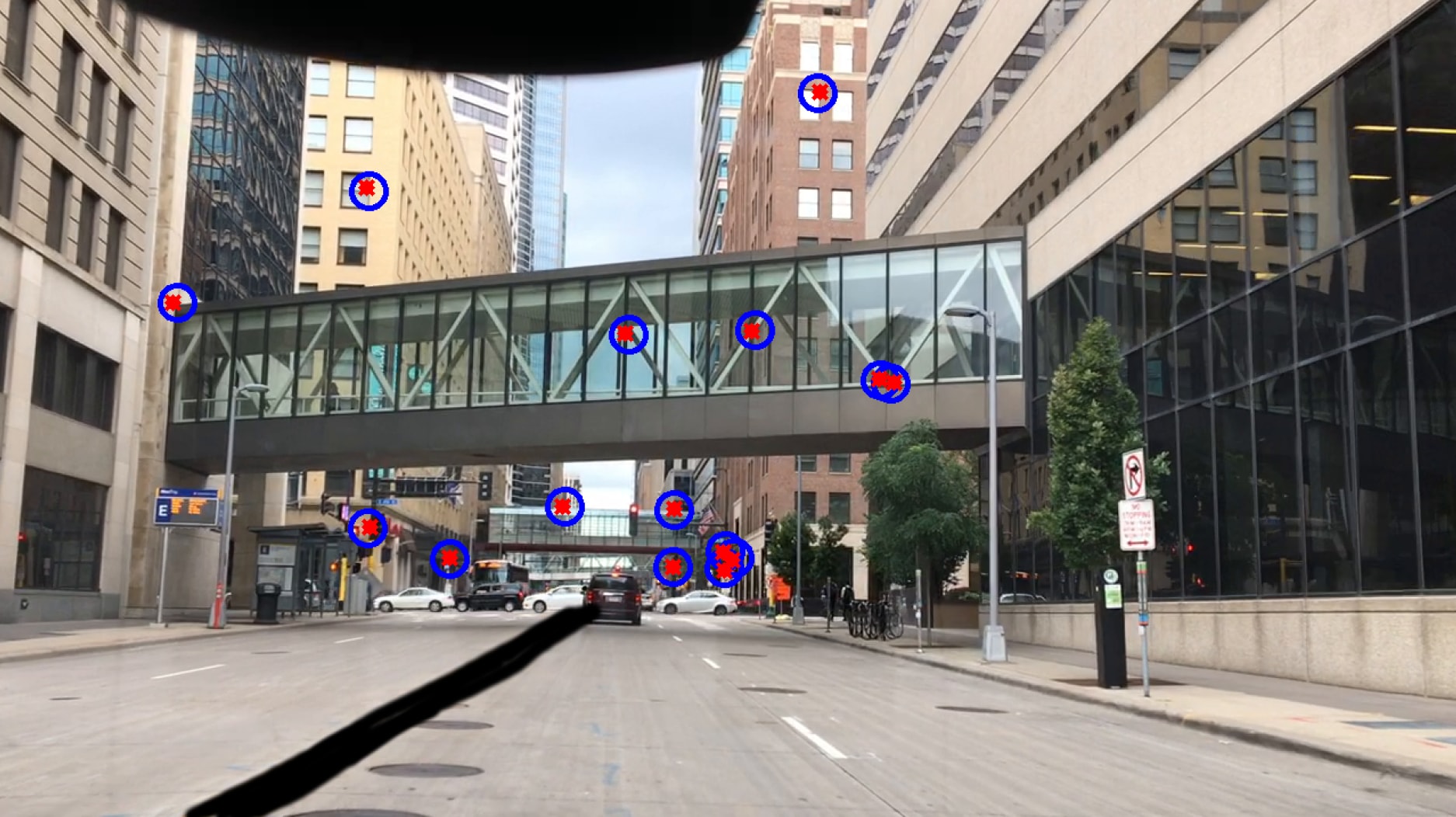}
   \caption{Projecting 3D feature points onto the current view. blue circle: matched feature point on current view; red marker: projected 3D feature point.}
   \label{fig:PnP_proj}
  \end{subfigure}
    ~ 
  \begin{subfigure}[t]{0.48\textwidth}
    \centering
    \includegraphics[width=\textwidth]{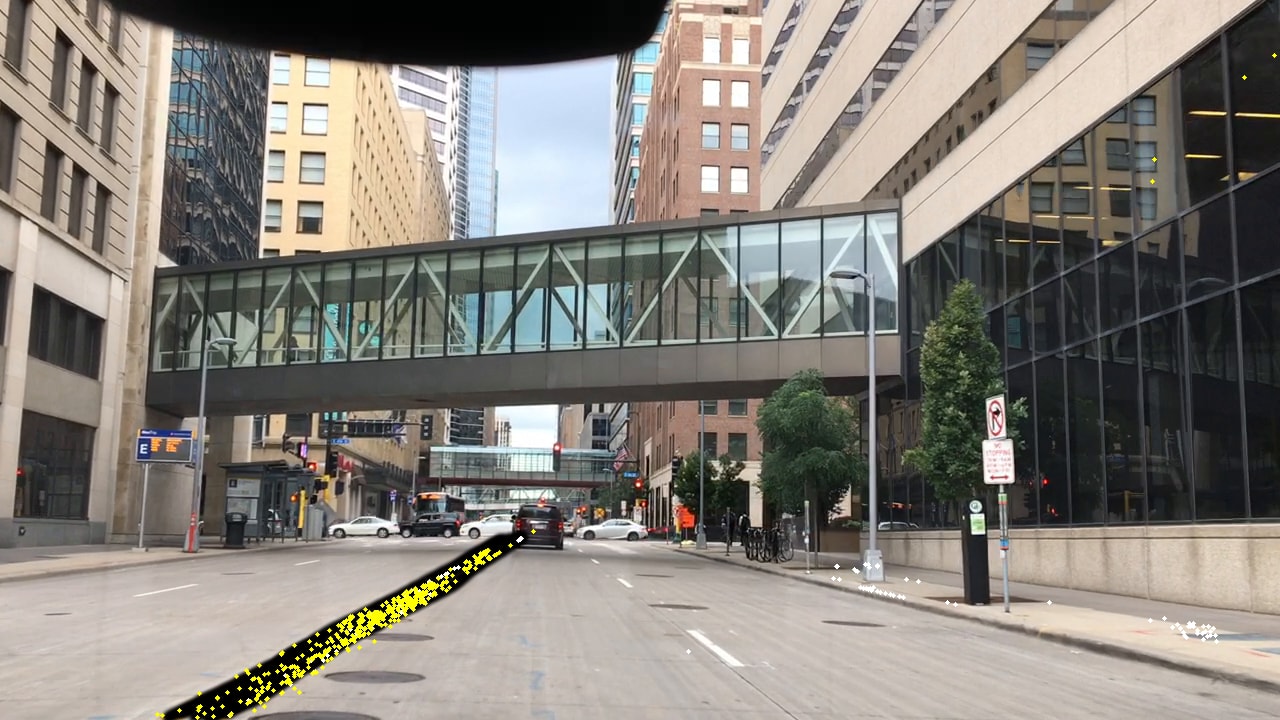}
    \caption{Final result.}
    \label{fig:result}
   \end{subfigure}
   \caption{Projecting lane markers onto the current view.}
\end{figure}

Next, we register the current camera view with the sparse 3D street model. We first match the current view with the database images so that the 3D street view is registered to the current view via 3D-2D correspondence, as illustrated in Figure~\ref{fig:PnP_inlier}. Then we use the 3D-2D correspondence to solve for the current camera pose. We get the rotation matrix:
\begin{equation*}
  \begin{bmatrix}
     0.999546 & -0.029875 & 0.004004 \\
     0.029743 &  0.999115 & 0.029731 \\
    -0.004889 & -0.029598 & 0.999550
  \end{bmatrix}
\end{equation*}
and translation vector:
\begin{equation*}
  \begin{bmatrix}
     0.068230 \\
    -0.261112 \\
    -2.135684
  \end{bmatrix}
\end{equation*}

To validate the rotation matrix and translation vector, we project the 3D feature points onto the current view, shown in Figure~\ref{fig:PnP_proj}. The average projection error is 2.19772 pixels.

Using the rotation matrix and translation vector obtained for the current camera, we are able to project the lane markers from the 3D street view onto the current view. The result is shown in Figure~\ref{fig:result}.

\begin{figure}[h!]
  \centering
    \includegraphics[width=0.48\textwidth]{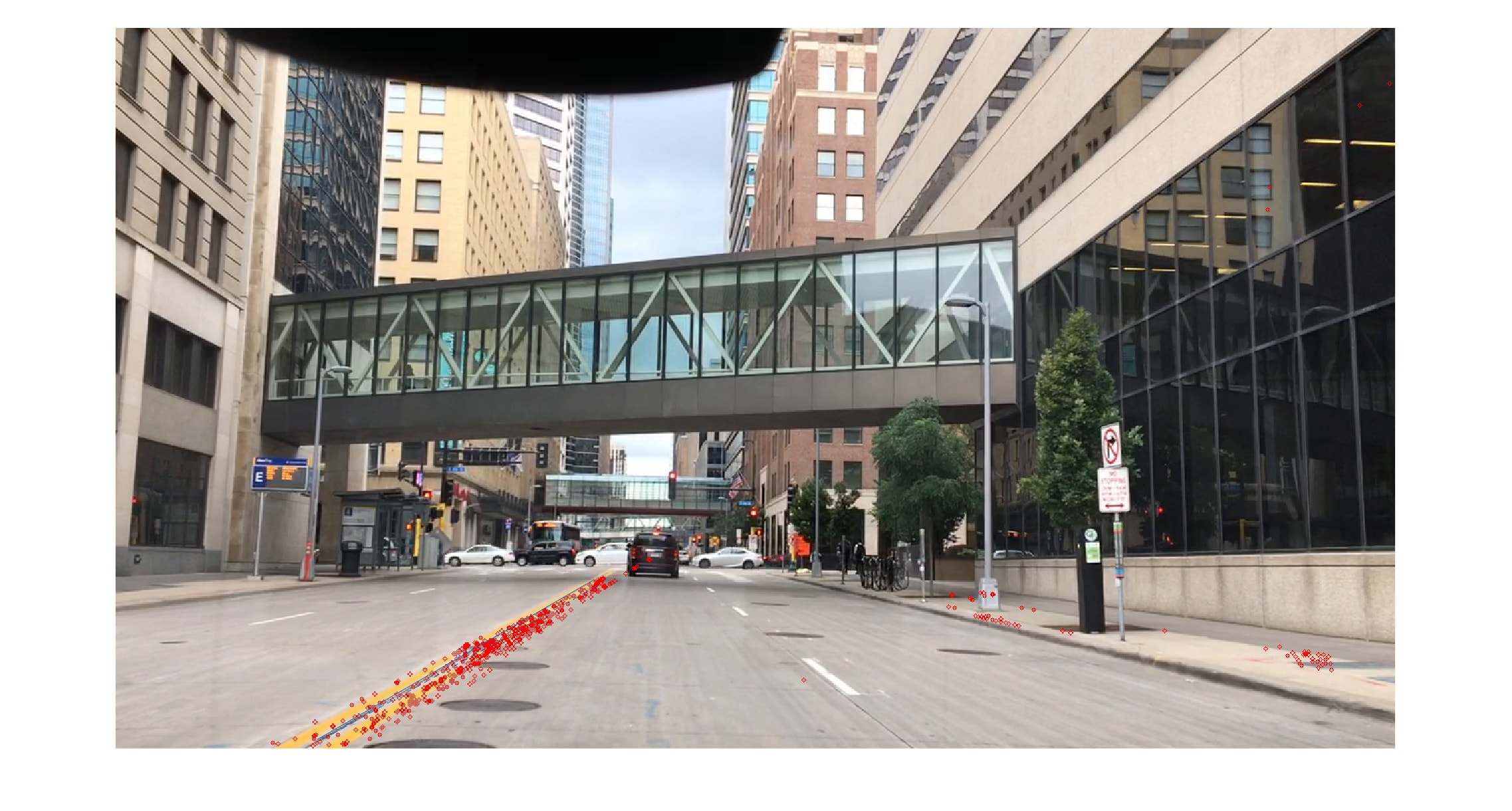}
   \caption{Difference between the projected line (dash line) and ground truth (solid line).}
   \label{fig:result_diff}
\end{figure}

To compare the projected pixels with ground truth, we manually select a line to optimally represent the projected pixels, shown as a dashed line in Figure~\ref{fig:result_diff}. At the same time, we select a line in the middle of lane marker as ground truth. We compare these two lines and find that the average offset between these two line segments is $4.803313$ pixels, where the image size is $1280 \times 720$ pixels.

\begin{figure}[h!]
  \centering
   \begin{subfigure}[t]{0.23\textwidth}
     \centering
     \includegraphics[width=\textwidth]{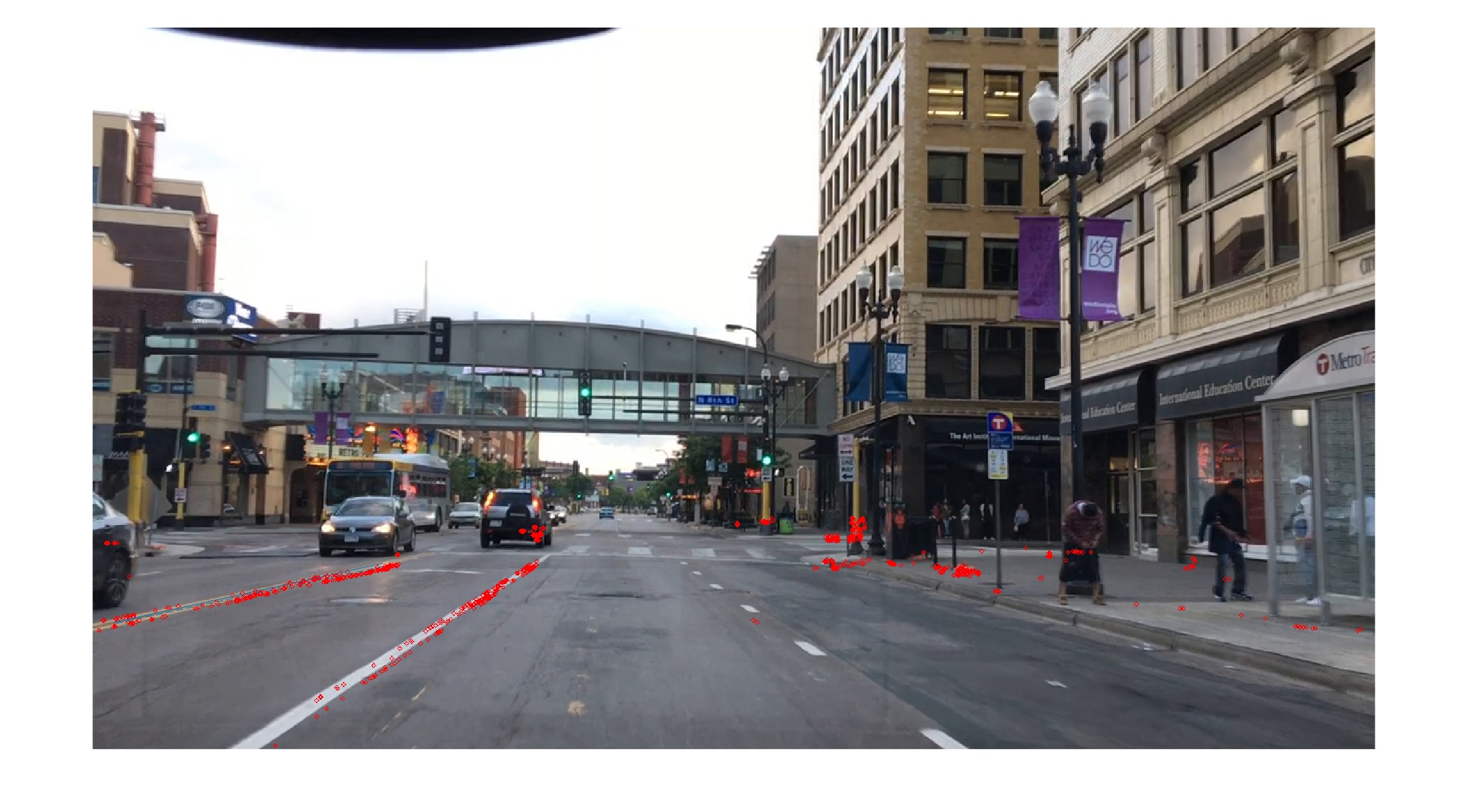}
     \caption{Average offset: 2.375941 pixels.}
   \end{subfigure}
   ~
   \begin{subfigure}[t]{0.23\textwidth}
     \centering
     \includegraphics[width=\textwidth]{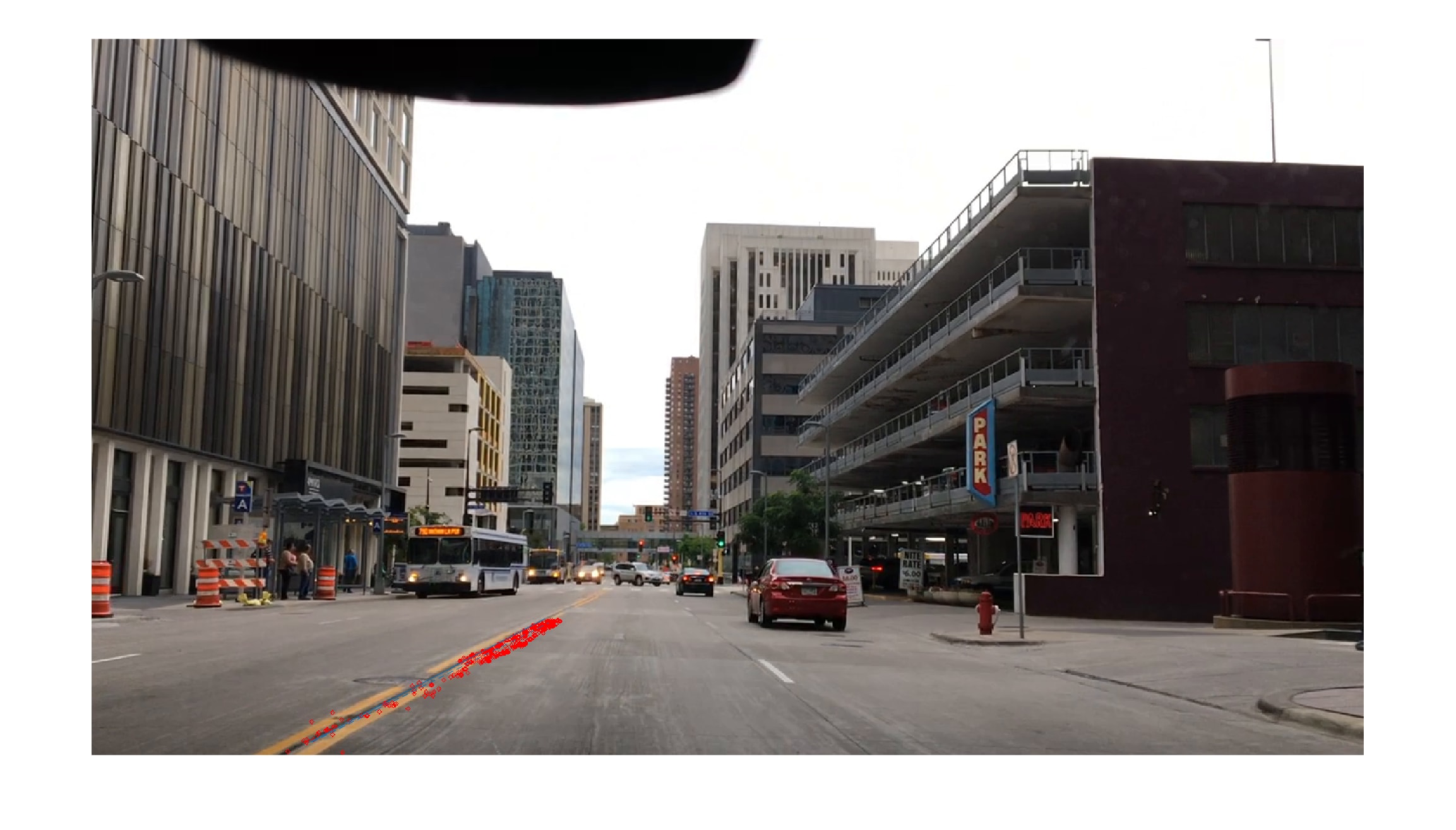}
     \caption{Average offset: 3.128704 pixels.}
   \end{subfigure}
   ~
  \begin{subfigure}[t]{0.23\textwidth}
    \centering
    \includegraphics[width=\textwidth]{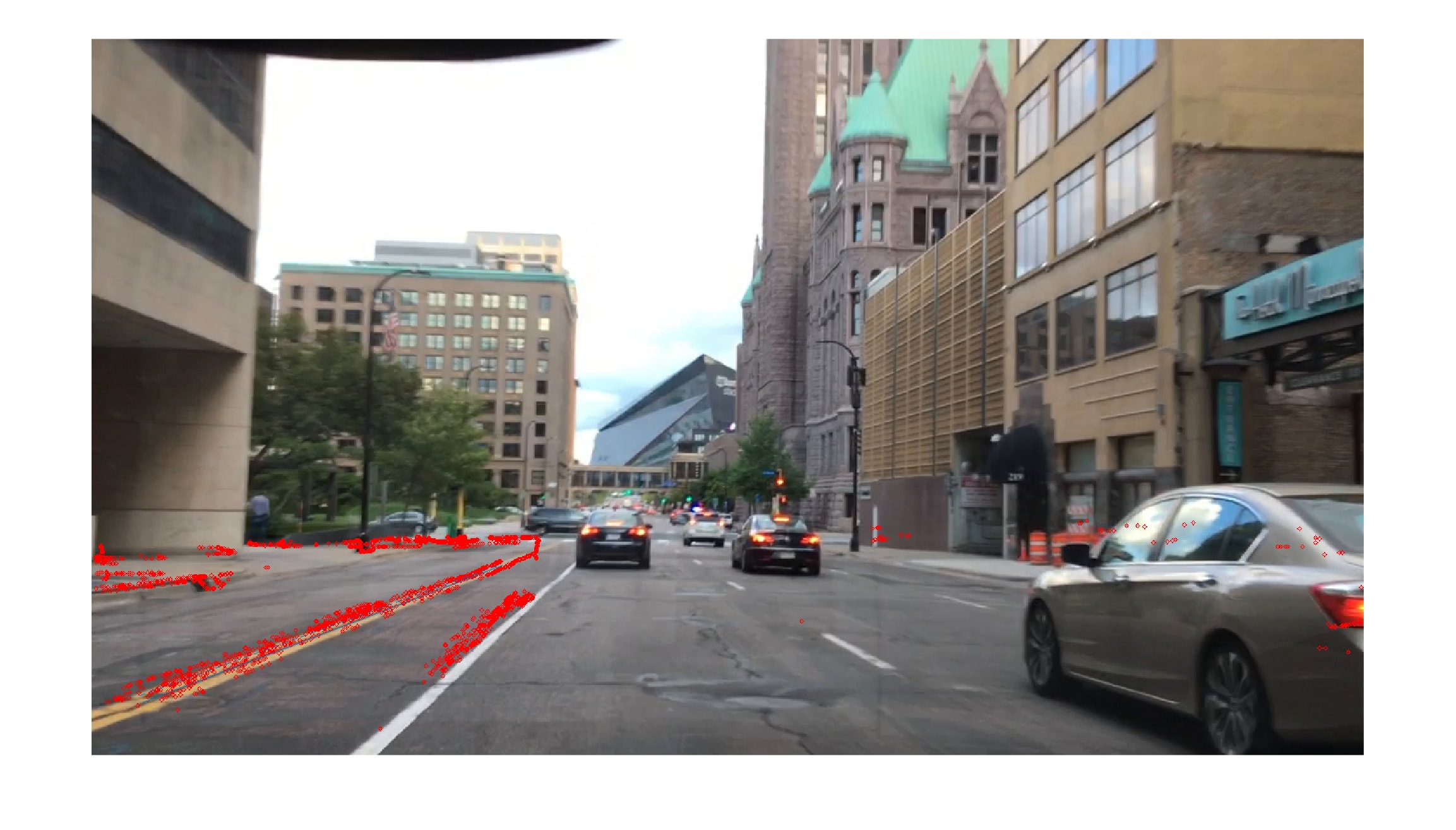}
    \caption{Average offset: 4.581539 pixels.}
   \end{subfigure}
    ~
   \begin{subfigure}[t]{0.23\textwidth}
     \centering
     \includegraphics[width=\textwidth]{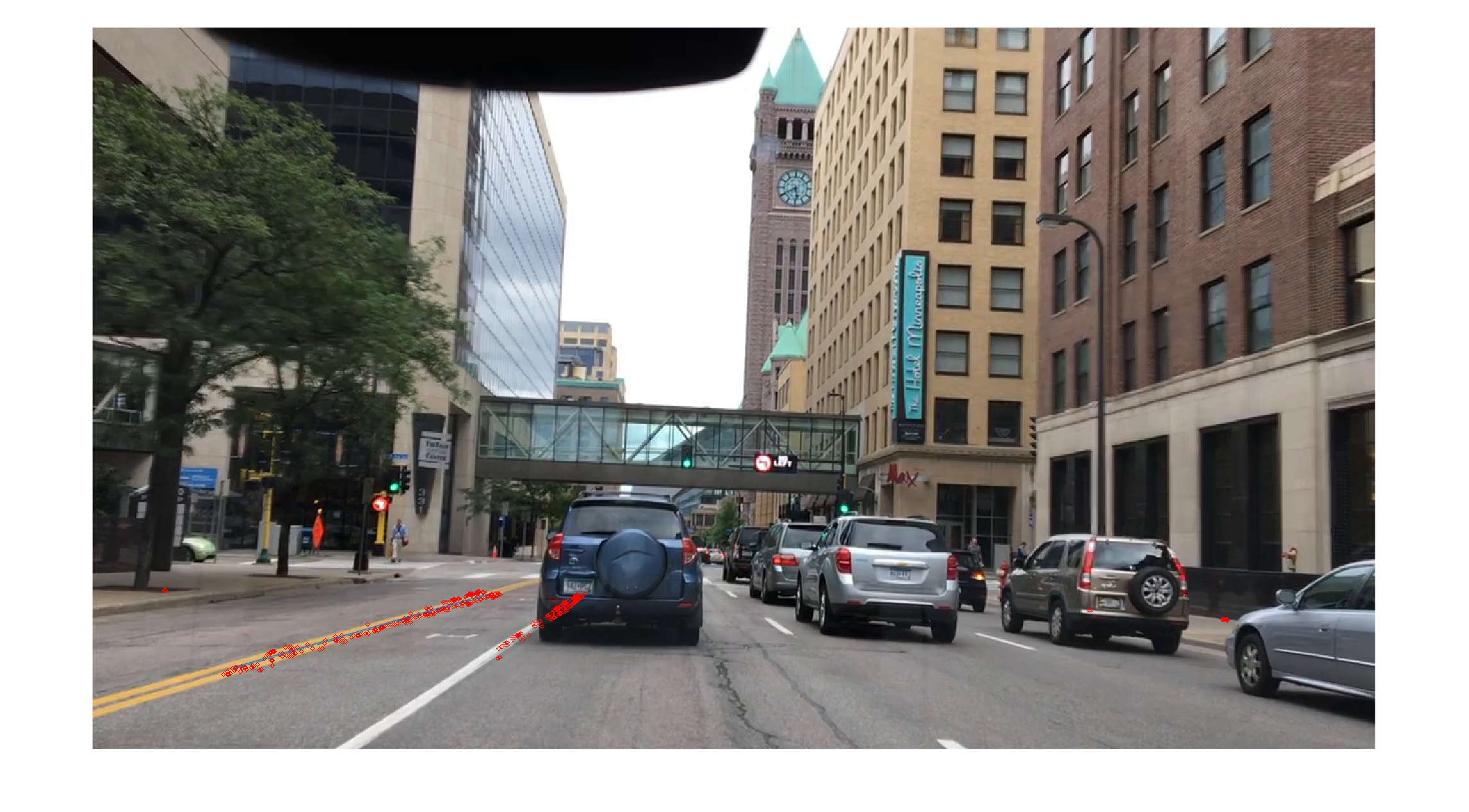}
     \caption{Average offset: 1.441206 pixels.}
   \end{subfigure}
   ~
   \begin{subfigure}[t]{0.23\textwidth}
     \centering
     \includegraphics[width=\textwidth]{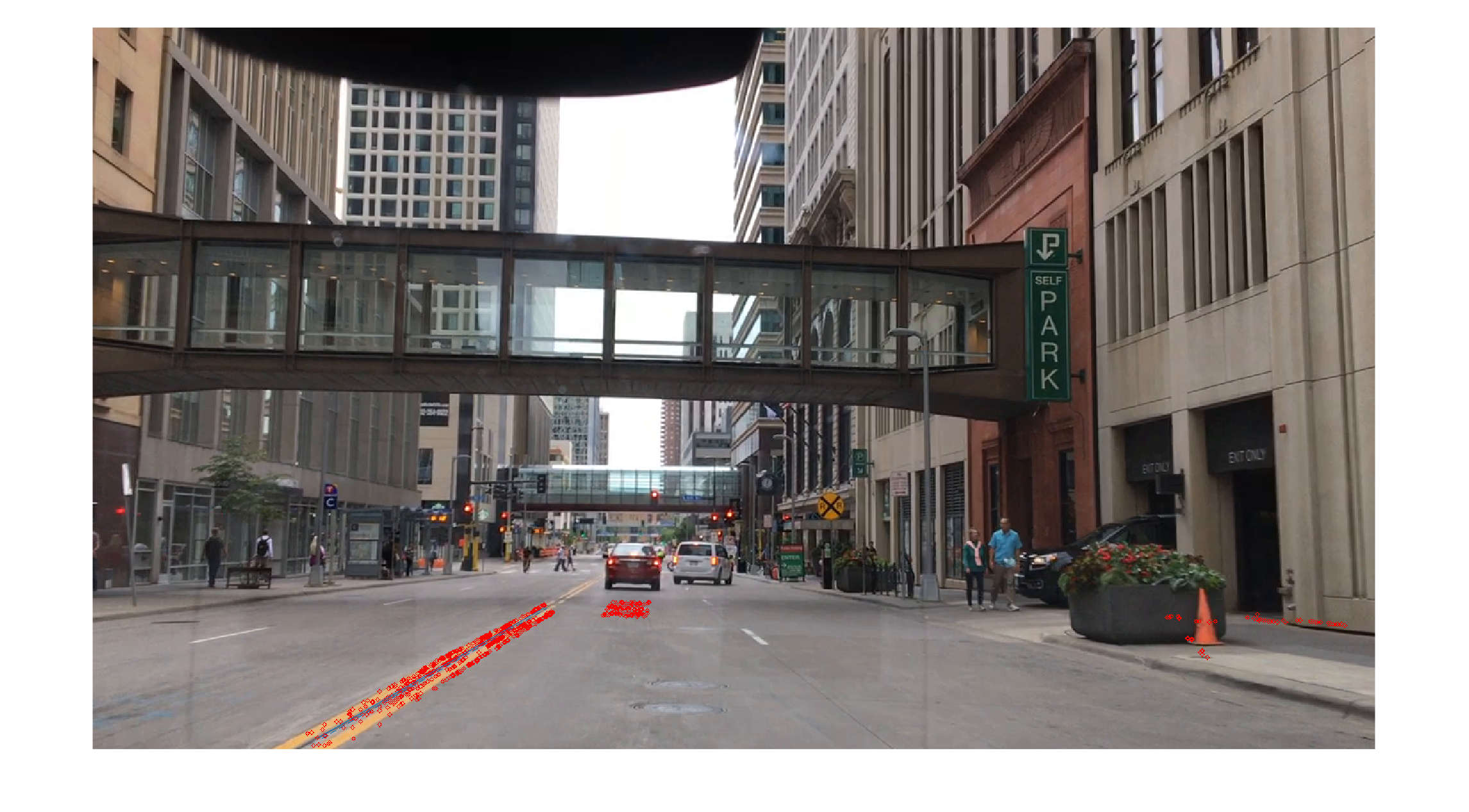}
     \caption{Average offset: 2.111444 pixels.}
   \end{subfigure}
    ~
   \begin{subfigure}[t]{0.23\textwidth}
     \centering
     \includegraphics[width=\textwidth]{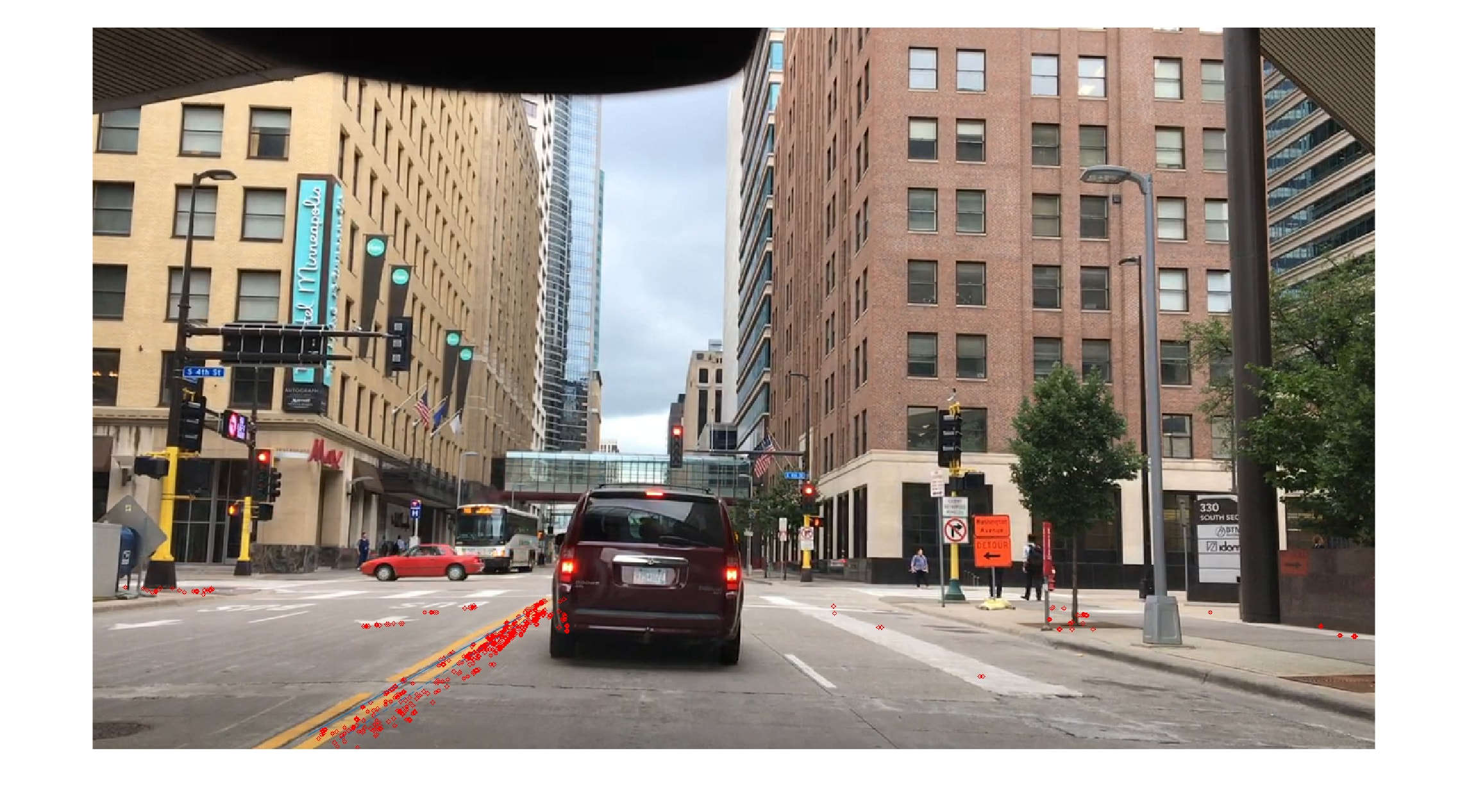}
     \caption{Average offset: 11.782807 pixels.}
   \end{subfigure}
   \caption{
   \label{fig:other_test_cases}%
   Other test cases, image size of $1280\times720$ pixels.
   }
\end{figure}

SafeDrive has been tested on a number of other scenarios, as illustrated in Figure~\ref{fig:other_test_cases}. It is important to note that SafeDrive is able to robustly locate lanes across a variety of scenes without any parameter tweaking, which is essential for autonomous operation.

\begin{figure}[h!]
  \centering
   \begin{subfigure}[t]{0.48\textwidth}
     \centering
    \includegraphics[width=\textwidth]{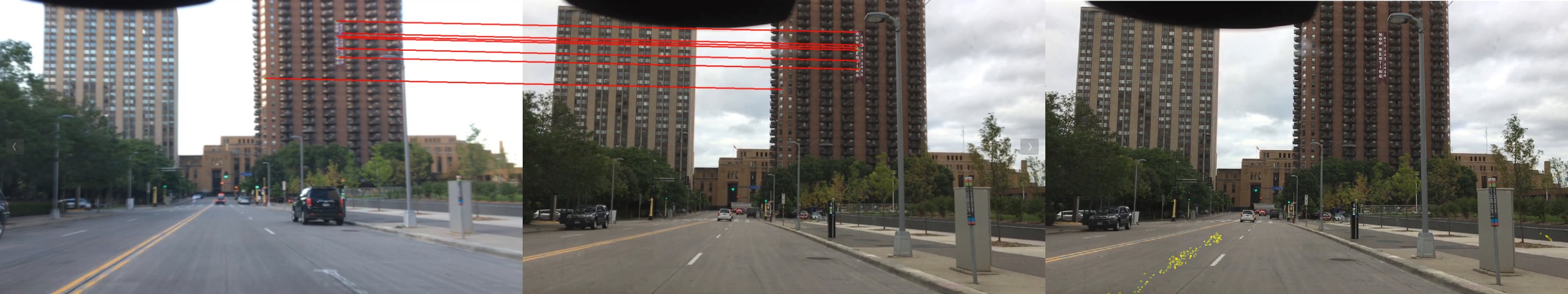}
   \caption{Large offset(72.312232 pixels) when 3D features are far away from current camera.}
   \label{fig:fail_case_1}
   \end{subfigure}
   ~
   \begin{subfigure}[t]{0.48\textwidth}
     \centering
    \includegraphics[width=\textwidth]{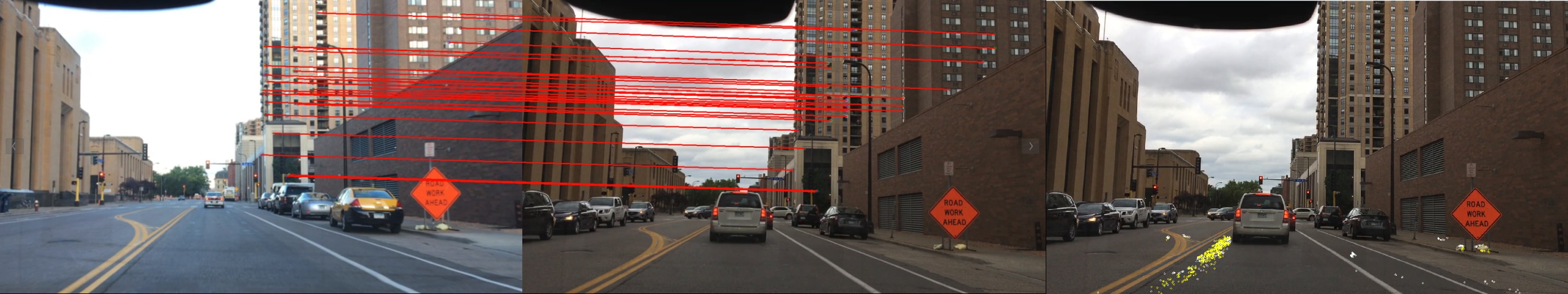}
   \caption{Large offset(36.181538 pixels) when most 3D features are on same side of the road.}
   \label{fig:fail_case_2}
   \end{subfigure}
   \caption{Instances where SafeDrive fails to accurately locate lanes markers.}
   \label{fig:fail_test_cases}
\end{figure}

However, when the current camera is far away from most 3D feature points in the street view, \emph{e.g.}, Figure~\ref{fig:fail_case_1}, or when 3D features are not equally spread, such as Figure~\ref{fig:fail_case_2}, performance of SafeDrive can degrade significantly.

\begin{figure}[h!]
  \centering
    \includegraphics[width=0.35\textwidth]{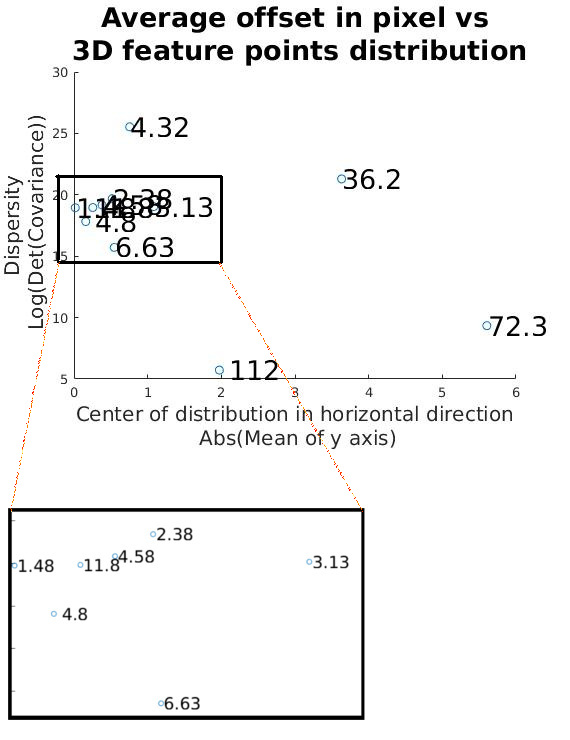}
   \caption{Average offset against 3D feature points distribution. The zoomed area shows low pixel offset from the ground truth, \emph{i.e.}, better accuracy, for high feature-point dispersity (widely spread) and low center of horizontal distribution (evenly distributed across the center vertical).}
   \label{fig:analysis}
\end{figure}

\begin{figure}[h!]
  \centering
   \begin{subfigure}[t]{0.48\textwidth}
     \centering
    \includegraphics[width=\textwidth]{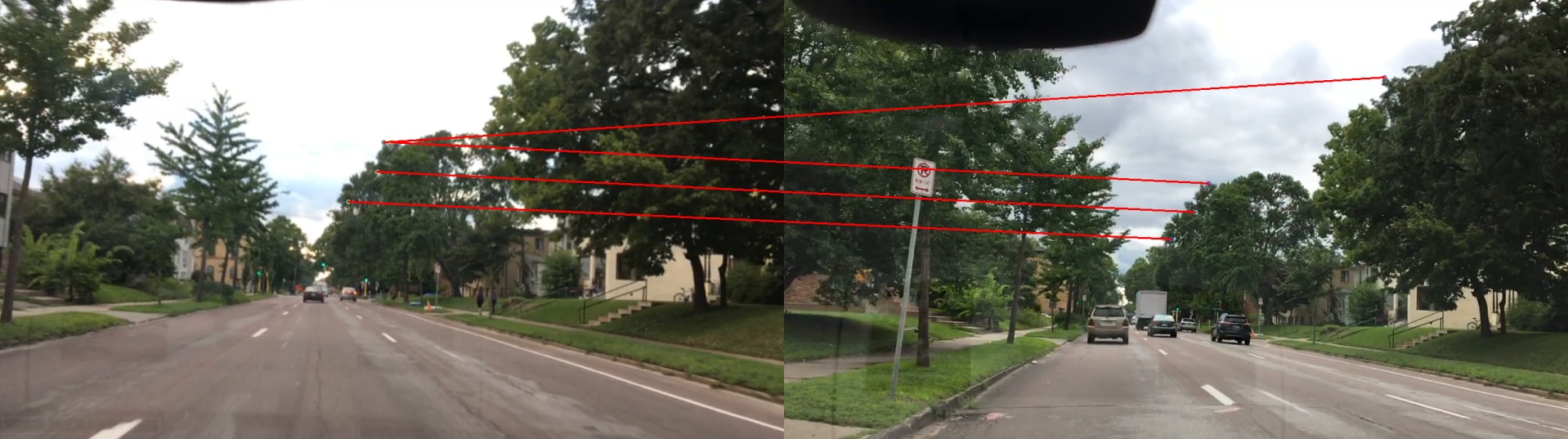}
   \caption{Feature correspondence on a rural road.}
   \label{fig:rural_match}
   \end{subfigure}
   ~
   \begin{subfigure}[t]{0.23\textwidth}
     \centering
    \includegraphics[width=\textwidth]{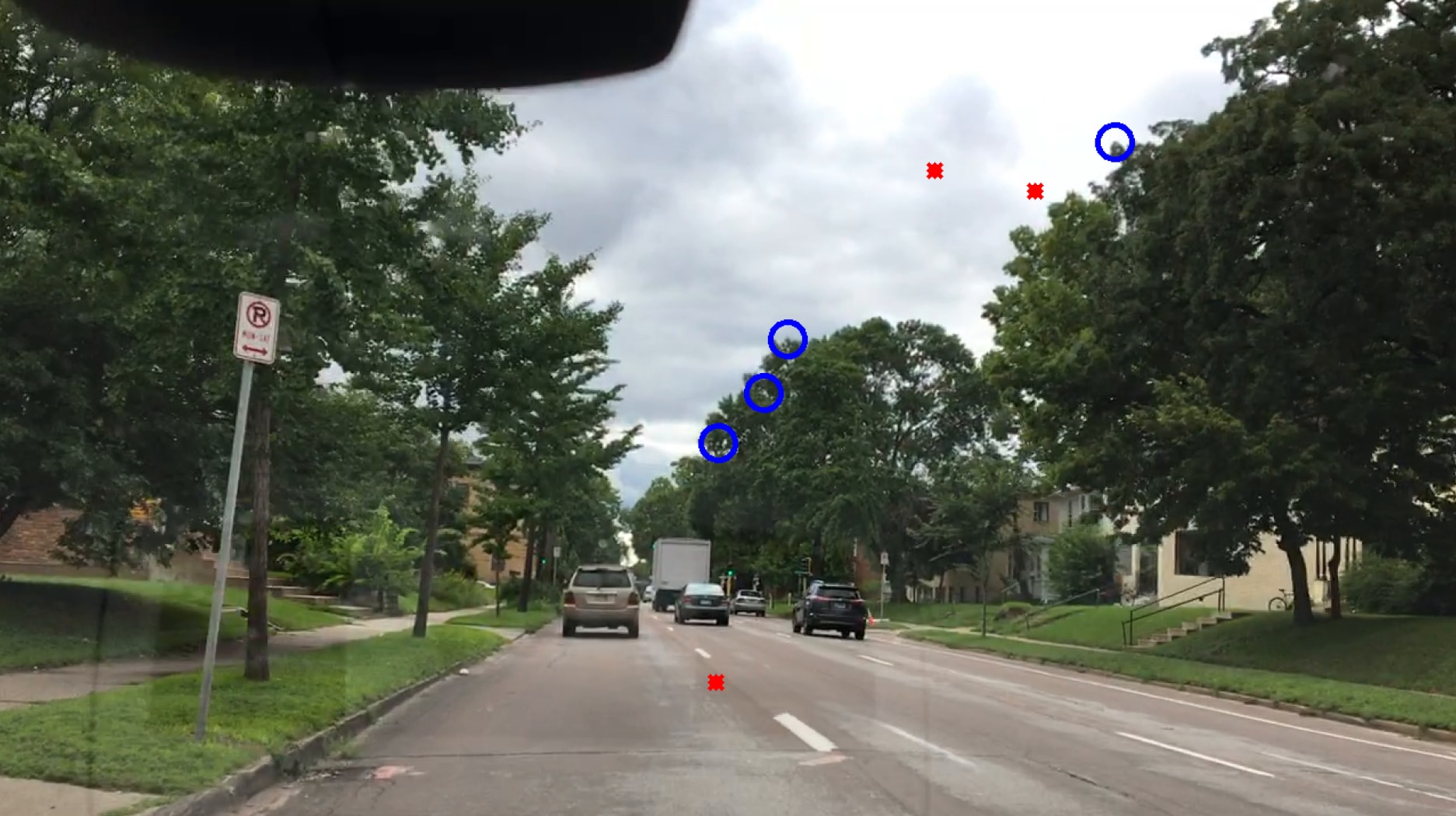}
   \caption{3D feature points projected onto the current view. blue circle: matched feature point on current view; red marker: projected 3D feature point.}
   \label{fig:rural_proj}
   \end{subfigure}
   ~
   \begin{subfigure}[t]{0.23\textwidth}
     \centering
    \includegraphics[width=\textwidth]{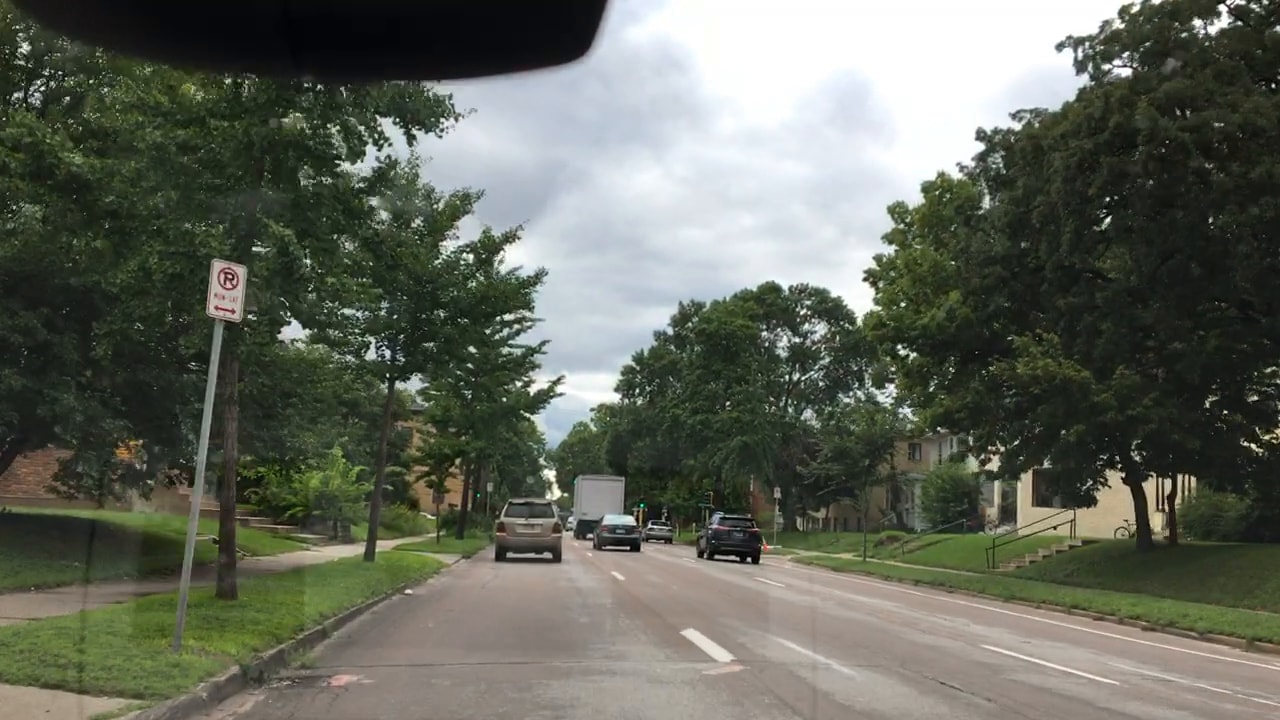}
   \caption{No visible projected lane marker, because projection matrix is wrong, indicated by the large feature projection error in Figure~\ref{fig:rural_proj}.}
   \label{fig:rural_res}
   \end{subfigure}
   \caption{A test case on a rural road-- SafeDrive accuracy suffers from feature sparsity.}
   \label{fig:rural_test_case}%
\end{figure}

Figure~\ref{fig:analysis} shows the effect of spatial distribution of 3D feature points on the accuracy of SafeDrive. For accurate lane marker projection, 3D feature points have to be distributed equally on both sides of the road, and be widely spread. Consequently, SafeDrive works better in urban areas, where feature points are usually more abundant. In rural areas, however, feature correspondences are neither enough or equally spread, causing performance degradation, as seen in Figure~\ref{fig:rural_test_case}. This shortcoming is currently being addressed by applying convolutional and generative models for feature selection.

\paragraph*{Performance}
SafeDrive has been implemented in C++ using OpenCV 2.4 bindings, and on an Intel\texttrademark{} Core i7 6700 CPU running Ubuntu 16.04, it takes $26.91$ seconds to run 12 test cases end-to-end (\emph{i.e.}, $2.24$ seconds per case) with a maximum of $2000$ feature points per image pair. We expect further improvements in speed once optimized.
\section{Conclusions}
We have presented an algorithm for enhancing lane marker appearance under poor road visibility, with the goal to improve visual lane detection for autonomous and assisted driving. This approach leverages the availability of alternate imagery at the current driving location and the ability to perform lane detection in such imagery, eventually projecting the lanes onto the original camera image. With sufficiently robust visual lane-finding algorithms, accurate pose detection, and robust methods to relate the past image with the live frame, we believe this algorithm can significantly improve driver safety. The ultimate goal for our work is to create an affordable system, and simultaneously improve the quality of autonomous transportation and occupant safety in road-going vehicles. Ongoing research is focusing on improved lane marker pixels matching, compressed data handling and optimization for enhanced performance, and extensive testing on data collected from a diverse range of geographic locations. 


\bibliographystyle{plain}
\small{
\bibliography{citation,allbibs}
}
\end{document}